\documentclass{article}

     \PassOptionsToPackage{numbers, compress}{natbib}
\usepackage[final]{neurips_2019}

\usepackage[utf8]{inputenc} % allow utf-8 input
\usepackage[T1]{fontenc}    % use 8-bit T1 fonts
\usepackage{hyperref}       % hyperlinks
\usepackage{url}            % simple URL typesetting
\usepackage{nicefrac}       % compact symbols for 1/2, etc.
\usepackage{microtype}      % microtypography
\usepackage{subfigure}
\usepackage{booktabs} % for professional tables
\usepackage{times}
\usepackage{amssymb,amsmath,amsfonts}
\usepackage{graphicx}   % need for figures
\usepackage{verbatim}   % useful for program listings
\usepackage{color}      % use if color is used in text
\usepackage{stmaryrd}
\usepackage{bbm}
\usepackage{multirow}
\usepackage{pifont}
\usepackage{textcomp}
\usepackage{dsfont}
\usepackage{graphicx}
\usepackage{algorithm}
\usepackage{algorithmic}

\newtheorem{theorem}{Theorem}
\newtheorem{definition}{Definition}
\newtheorem{lemma}{Lemma}

\newtheorem{corollary}{Corollary}
\newtheorem{remark}{Remark}

\newcommand{\bp}{\noindent{\emph{Proof}.}\ }
\newcommand{\ep}{\hfill $\Box$}

\newcommand{\BEAS}{\begin{eqnarray*}}
\newcommand{\EEAS}{\end{eqnarray*}}
\newcommand{\BEA}{\begin{eqnarray}}
\newcommand{\EEA}{\end{eqnarray}}
\newcommand{\BEQ}{\begin{equation}}
\newcommand{\EEQ}{\end{equation}}
\newcommand{\BIT}{\begin{itemize}}
\newcommand{\EIT}{\end{itemize}}
\newcommand{\BNUM}{\begin{enumerate}}
\newcommand{\ENUM}{\end{enumerate}}
\newcommand{\beq}{\begin{equation}}
\newcommand{\eeq}{\end{equation}}
\newcommand{\beqa}{\begin{eqnarray}}
\newcommand{\eeqa}{\end{eqnarray}}
\newcommand{\beqan}{\begin{eqnarray*}}
\newcommand{\eeqan}{\end{eqnarray*}}
\newcommand{\bealn}{\begin{align*}}
\newcommand{\eealn}{\end{align*}}
\newcommand{\al}[1]{ \begin{align} #1  \end{align}}

\newcommand{\als}[1]{ \begin{align*} #1  \end{align*}}

\newcommand{\sk}{\nonumber\\}

\newcommand{\el}{\end{flushleft}}
\newcommand{\bl}{\begin{flushleft}}

\newcommand{\argmax}{\arg\!\max}

\newcommand{\bI}{\mathbb{I}}

\newcommand{\cA}{\mathcal{A}}

\newcommand{\cD}{\mathcal{D}}

\newcommand{\cF}{\mathcal{F}}

\newcommand{\cO}{\mathcal{O}}

\newcommand{\cS}{\mathcal{S}}

\newcommand{\Real}{\mathbb{R}}
\newcommand{\Nat}{\mathbb{N}}

\newcommand{\indic}{\mathbb{I}}

\renewcommand{\phi}{\varphi}
\renewcommand{\epsilon}{\varepsilon}

\def\NN{{\mathbb N}}
\def\EE{{\mathbb E}}
\def\PP{{\mathbb P}}
\def\RR{{\mathbb R}}

\def\Acal{{\mathcal A}}
\def\Pcal{{\mathcal P}}

\def\Ocal{{\mathcal O}}

\def\Scal{{\mathcal S}}

\newcommand{\Esp}{\mathbb{E}}

\usepackage[colorinlistoftodos, textwidth=4cm, shadow]{todonotes}

\newcommand{\BAMC}{{\color{red!50!black}\textbf{BA-MC}}}
\newcommand{\ncut}{{$n_{\text{cutoff}}$}}

\title{Learning Multiple Markov Chains \\via Adaptive Allocation}

\author{ Mohammad Sadegh Talebi \\
   SequeL Team, Inria Lille -- Nord Europe\\
  \texttt{sadegh.talebi@inria.fr} \\
  \And
   Odalric-Ambrym Maillard \\
  SequeL Team, Inria Lille -- Nord Europe\\
   \texttt{sadegh.talebi@inria.fr} \\
  }

\begin{document}

\maketitle

\begin{abstract}
We study the problem of learning the transition matrices of a set of Markov chains from a single stream of observations on each chain. We assume that the Markov chains are ergodic but otherwise unknown. The learner can sample Markov chains sequentially to observe their states. The goal of the learner is to sequentially select various chains to learn transition matrices uniformly well with respect to some loss function. We introduce a notion of loss that naturally extends the squared loss for learning distributions to the case of Markov chains, and further characterize the notion of being \emph{uniformly good} in all problem instances. We present a novel learning algorithm that efficiently balances \emph{exploration} and \emph{exploitation} intrinsic to this problem, without any prior knowledge of the chains. We provide finite-sample PAC-type guarantees on the performance of the algorithm. Further, we show that our algorithm asymptotically attains an optimal loss.
\end{abstract}

\section{Introduction}\label{sec:intro}
We study a \emph{variant} of the following sequential adaptive allocation problem: %  :
A learner is given a set of $K$ arms, where to each arm $k\!\in\![K]$, an unknown real-valued distributions $\nu_k$ with mean $\mu_k$ and variance $\sigma^2_k\!>\!0$ is associated. At each round $t\!\in\!\Nat$, the learner	must select an arm $k_t\!\in\![K]$, and receives a sample drawn from $\nu_k$. Given a total budget of $n$ pulls, the objective is to estimate the expected values $(\mu_k)_{k\in[K]}$ of all distributions uniformly well. The quality of estimation in this problem is classically measured through \emph{expected} quadratic estimation error,
$\Esp[(\mu_k\!-\!\hat \mu_{k,n})^2]$ for the empirical mean estimate $\hat \mu_{k,n}$ built with the $T_{k,n}\!=\!\sum_{t=1}^n \indic\{k \!=\!k_t\}$ many samples received from $\nu_k$ at time $n$, and the performance of an allocation strategy is the maximal error, $\max_{k\in[K]} \Esp[(\mu_k\!-\!\hat \mu_{k,n})^2]$. Using ideas from the Multi-Armed Bandit (MAB) literature, previous works (e.g.,  \cite{antos2008active,carpentier2011upper}) have provided optimistic sampling strategies with near-optimal performance guarantees for this setup.

This generic adaptive allocation problem is related to several application problems arising in optimal experiment design \cite{fedorov1972theory,dror2008sequential}, active learning \cite{cohn1996active}, or Monte-Carlo methods \cite{etore2010adaptive}; we refer to \cite{antos2008active,antos2010active,carpentier2011upper,carpentier2015adaptive} and references therein for further motivation. We extend this line of work to the case where each process is a \emph{discrete Markov chain}, hence introducing the problem of \emph{active bandit learning of Markov chains}.
More precisely, we no longer assume that $(\nu_k)_k$ are real-valued distributions, but we study the case where each $\nu_k$ is a discrete Markov process over a state space $\cS\subset\Nat$.
The law of the observations $(X_{k,i})_{i\in\Nat}$ on arm (or chain) $k$ is given by $\nu_k(X_{k,1},\dots X_{k,n}) =  p_k(X_{k,1})\prod_{i=2}^n P_k(X_{k,i-1},X_{k,i})$, where
$p_k$ denotes the initial distribution of states, and  $P_k$ is the transition function of the Markov chain.  The goal of the learner is to learn the \emph{transition matrices} $(P_k)_{k\in[K]}$ uniformly well on the chains. Note that the chains are \emph{not} controlled (we only decide which chain to advance, not the states it transits to).

Before discussing the challenges of the extension to Markov chains,
let us give further comments on the performance measure considered in bandit allocation for real-valued distributions: Using the expected quadratic estimation error on each arm $k$ makes sense since when $T_{k,n}, k\in [K]$ are deterministic, it coincides with $\sigma^2_k/T_{k,n}$, thus suggesting to pull the distributions proportionally to $\sigma^2_k$. However, for a learning strategy, $T_{k,n}$  typically depends on all past observations.
The presented analyses in these series of works rely on Wald's second identity as the technical device, heavily relying on the use of a quadratic loss criterion, which prevents one from extending the approach therein to other distances.
Another peculiarity arising in working with expectations is the order of ``max'' and ``expectation'' operators. While it makes more sense to control  the \emph{expected value of the maximum}, the works cited above look at \emph{maximum of the expected value}, which is more in line with a pseudo-loss definition rather than the loss; actually in extensions of these works a pseudo-loss is considered instead of this performance measure. As we show, all of these difficulties can be avoided by resorting to a \emph{high probability} setup. Hence, in this paper, we deviate from using an \emph{expected} loss criterion, and rather use a high-probability control. We formally define our performance criterion in Section~\ref{sub:perfmeasure}.

\subsection{Related Work}\label{sub:related}
%
%\oam{Reshuffle!  Most relevant to least relevant}

On the one hand, our setup can be framed into  the line of works on active bandit allocation, considered for the estimation of reward distributions in MABs as introduced in \cite{antos2008active,antos2010active}, and further studied in \cite{carpentier2011upper,neufeld2014adaptive}. This has been extended to stratified sampling for Monte-Carlo methods in \cite{carpentier2012adaptive,carpentier2015adaptive}, or to  continuous mean functions in, e.g., \cite{carpentier2012online}. On the other hand, our extension from real-valued distributions to Markov chains can be framed into the rich literature on Markov chain estimation; see, e.g., \cite{billingsley1961statistical,kipnis1986central,haviv1984perturbation,welton2005estimation,craig2002estimation,meyn2012markov}.
This stream of works extends a wide range of results from the i.i.d.~case to the Markov case. These include, for instance, the law of large numbers for (functions of) state values \cite{meyn2012markov}, the central limit theorem for Markov sequences \cite{kipnis1986central} (see also \cite{meyn2012markov,rio2017asymptotic}), and Chernoff-type or Bernstein-type concentration inequalities for Markov sequences \cite{lezaud1998chernoff,paulin2015concentration}.
Note that the majority of these results are available for \emph{ergodic} Markov chains.

Another stream of research on Markov chains, which is more relevant to our work, investigates learning and estimation of the \emph{transition matrix} (as opposed to its full law); see, e.g., \cite{craig2002estimation,welton2005estimation,wolfer2019minimax,hao2018learning}.
Among the recent studies falling in this category, \cite{hao2018learning} investigates learning of the transition matrix with respect to a loss function induced by $f$-divergences in a \emph{minimax setup}, thus extending \cite{kamath2015learning} to the case of Markov chains.
\cite{wolfer2019minimax} derives a PAC-type bound for learning the transition matrix of an ergodic Markov chain with respect to the \emph{total variation} loss. It further provides a matching lower bound. Among the existing literature on learning Markov chains, to the best of our knowledge, \cite{wolfer2019minimax} is the closest to ours. There are however two aspects distinguishing our work: Firstly, the challenge in our problem resides in dealing with \emph{multiple} Markov chains, which is present neither in \cite{wolfer2019minimax} nor in the other studies cited above. Secondly, our notion of loss does not coincide with that considered in \cite{wolfer2019minimax}, and hence, the lower bound of \cite{wolfer2019minimax} does not apply to our case.

Among the results dealing with multiple chains, we may refer to learning in the Markovian bandits setup \cite{ortner2014regret,tekin2012online,dance2019optimal}. Most of these studies address the problem of reward maximization over a finite time horizon. We also mention that in a recent study, \cite{tarbouriech2019active} introduces the so-called active exploration in Markov decision processes, where the transition kernel is \emph{known}, and the goal is rather to learn the mean reward associated to various states. To the best of our knowledge, none of these works address the problem of learning the \emph{transition matrix}.
Last, as we target  high-probability performance bounds (as opposed to those holding in expectation), our approach is naturally linked to the Probably Approximately Correct (PAC) analysis.
%There is a large body of literature on estimation and learning of discrete distributions, and their extensions to Markov chains.
\cite{kearns1994learnability} provides one of the first PAC bounds for learning discrete distributions. Since then, the problem of learning discrete distributions has been  well studied; see, e.g., \cite{gamarnik2003extension,jiao2015minimax,kamath2015learning} and references therein. We refer to \cite{kamath2015learning} for a rather complete characterization of learning distribution in a \emph{minimax} setting under a big class of smooth loss functions. We remark that except for very few studies (e.g., \cite{gamarnik2003extension}), most of these results are provided for discrete distributions.

\subsection{Overview and Contributions}\label{sub:contrib}
Our contributions are the following:
(i) For the problem of learning Markov chains, we consider  a notion of loss function, which appropriately extends the loss function for learning distributions to the case of Markov chains. Our notion of loss is similar to that considered in \cite{hao2018learning} (we refer to Section \ref{sub:perfmeasure} for a comparison between our notion and the one in \cite{hao2018learning}). In contrast to existing works on similar bandit allocation problems, our loss function avoids technical difficulties faced when extending the squared loss function to this setup. We further characterize the notion of a ``uniformly good algorithm'' under the considered loss function for ergodic chains; (ii) We present an \emph{optimistic} algorithm, called \BAMC, for active learning of Markov chains, which is simple to implement and does not require any prior knowledge of the chains. To the best of our knowledge, this constitutes the first algorithm for active bandit allocation for learning Markov chains; (iii) We provide non-asymptotic PAC-type, and asymptotic bounds, on the loss incurred by \BAMC, indicating three regimes. In the first regime, which holds for any budget $n\ge 4K$, we present (in Theorem \ref{thm:loss_squared_generic}) a high-probability bound on the loss scaling as $\widetilde \Ocal(\frac{KS^2}{n})$, where $\widetilde \cO(\cdot)$ hides $\log(\log(n))$ factors. Here, $K$ and $S$ respectively denote the number of chains and the number of states in a given chain. This result holds for homogenous Markov chains. We then characterize a cut-off budget \ncut\ (in Theorem \ref{thm:loss_squared_suff_budget}) so that when $n\ge \text{\ncut}$, the loss behaves as $\widetilde \cO(\frac{\Lambda}{n} + \frac{C_0}{n^{3/2}})$, where $\Lambda = \sum_k \sum_{x,y} P_k(x,y)(1 - P_k(x,y))$ denotes the sum of variances of all states and all chains, and where $P_k$ denotes the transition probability of chain $k$. This latter bound constitutes the second regime, in view of the fact that $\frac{\Lambda}{n}$ equals the asymptotically optimal loss (see Section \ref{sec:static} for more details). Thus, this bound indicates that the \emph{pseudo-excess loss} incurred by the algorithm vanishes at a rate $C_0n^{-3/2}$ (we refer to Section \ref{sec:bounds} for a more precise definition). Furthermore, we carefully characterize the constant $C_0$. In particular, we discuss that  $C_0$ does not deteriorate with mixing times of the chains, which, we believe, is a strong feature of our algorithm. We also discuss how various properties of the chains, e.g., discrepancies between stationary distribution of various states a given chain, may impact the learning performance. Finally, we demonstrate a third regime, the asymptotic one, when the budget $n$ grows large, in which we show (in Theorem \ref{thm:loss_squared_asymp}) that the loss of \BAMC\
matches the asymptotically optimal loss $\frac{\Lambda}{n}$. All proofs are provided in the supplementary material.
%; (iv) Finally, we briefly discuss the case of restless chains and provide a heuristic way to tailor our algorithms to this case.

Markov chains have been successfully used for modeling a broad range of practical problems, and their success makes the studied problem in this paper relevant in practice. There are practical applications in reinforcement learning (e.g., active exploration in MDPs \cite{tarbouriech2019active}) and in rested Markov bandits (e.g., channel allocation in wireless communication systems where a given channel's state follows a Markov chain\footnote{For example, in the Gilbert-Elliott channels \cite{mushkin1989capacity}.}), for which we believe our contributions could serve as a technical tool.

\section{Preliminaries and Problem Statement}\label{sec:prelim}
%In this section, we briefly provide the required background material on Markov chains.

\subsection{Preliminaries}
Before describing our model, we recall some preliminaries on Markov chains; these are standard definitions and results, and can be found in, e.g., \cite{norris1998markov,levin2009markov}. Consider a Markov chain defined on a finite state space $\Scal$ with cardinality $S$. Let $\Pcal_{\Scal}$ denote the collection of all row-stochastic matrices over $\Scal$. The Markov chain is specified by its transition matrix $P\in \Pcal_{\Scal}$ and its initial distribution $p$: For all $x,y\in \Scal$, $P(x,y)$  denotes the probability of transition to $y$ if the current state is $x$. In what follows, we may refer to a chain by just referring to its transition matrix.

We recall that a Markov chain $P$ is \emph{ergodic} if $P^m>0$ (entry-wise) for some $m\in \NN$. If $P$ is ergodic, then it has a unique \emph{stationary distribution} $\pi$ satisfying $\pi = \pi P$. Moreover $\underline{\pi} := \min_{x\in \cS} \pi(x) >0$. A chain $P$ is said to be \emph{reversible} if its stationary distribution $\pi$ satisfies \emph{detailed balance equations}: For all $x,y\in \Scal$, $\pi(x)P(x,y) = \pi(y)P(y,x)$. Otherwise, $P$ is called \emph{non-reversible}. For a Markov chain $P$, the largest eigenvalue is $\lambda_1(P)=1$ (with multiplicity one).
In a \emph{reversible} chain $P$, all eigenvalues belong to $(-1,1]$.  We define the \emph{absolute spectral gap} of a reversible chain $P$ as $\gamma(P) = 1 - \lambda_\star(P)$, where $\lambda_\star(P)$ denotes the second largest (in absolute value) eigenvalue of $P$.
If $P$ is reversible, the absolute spectral gap $\gamma(P)$ controls the convergence rate of the state distributions of the chain towards the stationary distribution $\pi$. If $P$ is \emph{non-reversible}, the convergence rate is determined by the \emph{pseudo-spectral gap} as introduced in \cite{paulin2015concentration} as follows.  Define $P^\star$ as: $P^\star(x,y) := \pi(y)P(y,x)/\pi(x)$ for all $x,y\in \Scal$. Then, the pseudo-spectral gap $\gamma_{\textsf{ps}}(P)$ is defined as:
$
\gamma_{\textsf{ps}}(P) := \max_{\ell\ge 1} \frac{\gamma((P^\star)^\ell P^\ell)}{\ell}.
$

%Let us denote by $\hat\pi_n$ the empirical stationary distribution of $P$ after $n$ steps. If the chain is reversible, then the convergence rate of $\hat\pi_n$ to $\pi$ is controlled by the absolute spectral gap $\gamma$. On the other hand,

\subsection{Model and Problem Statement}
We are now ready to describe our model.  We consider a learner interacting with a finite set of Markov chains indexed by $k\in [K]:=\{1,2,\ldots,K\}$. For  ease of presentation, we assume that all Markov chains are defined on the same state space\footnote{Our algorithm and results are straightforwardly extended to the case where the Markov chains are defined on different state spaces.} $\Scal$ with cardinality $S$. The Markov chain $k$, or for short chain $k$, is specified by its transition matrix $P_k\in \Pcal_{\Scal}$. In this work, we assume that all Markov chains are ergodic, which implies that any chain $k$ admits a unique stationary distribution, which we denote by $\pi_k$. Moreover, the minimal element of $\pi_k$ is bounded away from zero: $\underline \pi_k:= \min_{x\in \cS} \pi_k(x)>0$. The initial distributions of the chains are assumed to be arbitrary. Further, we let $\gamma_k:=\gamma(P_k)$ to denote the absolute spectral gap of chain $k$ if $k$ is reversible; otherwise, we define the pseudo-spectral gap of $k$ by $\gamma_{\textsf{ps},k}:=\gamma_{\textsf{ps}}(P_k)$.

A related quantity in our results is the \emph{Gini index} of the various states. %For a row probability vector $u$, define the function
For a chain $k$, the \emph{Gini index} for state $x\in \Scal$ is defined as
$$
G_k(x):= \sum_{y\in \Scal} P_k(x,y)(1- P_k(x,y)).
$$
Note that  $G_k(x) \le 1-\frac{1}{S}$. This upper bound is verified by the fact that the maximal value of $G_k(x)$ is achieved when $P_k(x,y)=\frac{1}{S}$ for all $y\in \Scal$ (in view of the concavity of $z\mapsto \sum_{x\in \cS} z(x)(1-z(x))$).
 In this work, we assume that for all $k$, $\sum_{x\in \Scal} G_k(x)>0$.\footnote{We remark that there exist chains  with $\sum_x G_k(x) = 0$. In view of the definition of the Gini index, such chains are necessarily deterministic (or degenerate), namely their transition matrices belong to $\{0,1\}^{S\times S}$. One example is a deterministic cycle with $S$ nodes. We note that such chains may fail to satisfy irreducibility or aperiodicity.}
Another related quantity in our results is the sum (over states) of inverse stationary distributions: For a chain $k$, we define $H_k:=\sum_{x\in \Scal} \pi_k(x)^{-1}$. Note that $S^2\le H_k \le S\underline{\pi}_k^{-1}$. The quantity $H_k$ reflects the discrepancy between individual elements of $\pi_k$.  %We finally remark that both $G_k(x)$ and $H_k$ are problem dependent

\paragraph{The online learning problem.}
The learner wishes to design a sequential allocation strategy to adaptively sample various Markov chains so that all transition matrices are learnt uniformly well.
The game proceeds as follows: Initially all chains are assumed to be non-stationary with arbitrary initial distributions chosen by the environment. At each step $t\ge 1$, the learner samples a chain $k_t$, based on the past decisions and the observed states, and observes the state $X_{k_t,t}$. The state of $k_t$ evolves according to $P_{k_t}$. The state of chains $k\ne k_t$ does not change: $X_{k,t}=X_{k,t-1}$ for all $k\ne k_t$.

We introduce the following notations: Let $T_{k,t}$ denote the number of times chain $k$ is selected by the learner up to time $t$: $T_{k,t}:=\sum_{t'=1}^t \bI\{k_{t'} = k\}$, where $\bI\{\cdot\}$ denotes the indicator function. Likewise, we let $T_{k,x,t}$ represent the number of observations  of chain $k$, up to time $t$, when the chain was in state $x$:
$
T_{k,x,t}:=\sum_{t'=1}^t \bI\{k_{t'} = k, X_{k,t'} = x\}.
$
Further, we note that the learner only controls $T_{k,t}$ (or equivalently, $\sum_{x} T_{k,x,t}$), but not the number of visits to individual states. At each step $t$, the learner maintains empirical estimates of the stationary distributions, and estimates transition probabilities of various chains based on the observations gathered up to $t$. We define the empirical stationary distribution of chain $k$ at time $t$ as
$
\hat\pi_{k,t}(x):=T_{k,x,t}/T_{k,t}$ for all $x\in \Scal.
$
For chain $k$, we maintain the following \emph{smoothed estimation} of transition probabilities:
\al{
\label{eq:P_estimator_alpha}
\widehat P_{k,t}(x,y):=\frac{\alpha + \sum_{t'=2}^t \bI\{X_{k, t'-1}=x, X_{k, t'}=y\}}{\alpha S + T_{k,x,t}}\, , \quad \forall x,y\in \Scal,
}
where $\alpha$ is a positive constant. In the literature, the case of $\alpha = \frac{1}{S}$ is usually referred to as the \emph{Laplace-smoothed} estimator.
The learner is given a budget of $n$ samples, and her goal is to obtain an accurate estimation of transition matrices of the Markov chains. The accuracy of the estimation is determined by some notion of loss, which will be discussed later. The learner adaptively selects various chains so that the minimal loss is achieved.

%Given a budget of $n$ samples, for an algorithm $\Acal$, we define the following loss functions:
%\als{
%L_{\mathrm{TV}, \mathcal A,n}&:= \max_{k\in [K]} L_{\mathrm{TV},k,n}\, , \\
%L_{\mathrm{F}, \mathcal A,n}&:= \max_{k\in [K]} L_{\mathrm{F},k,n}\, .
%}
%\als{
%\PP\left( \max_k \max_{x,y} |P_k(x,y) - \widehat{P}_{k,n}(x,y)| \ge \epsilon \right) &\le \delta \\
%\PP\left( \max_k \vertiii{P_k - \widehat P_{k,n}}_{\mathrm{F}} \ge \epsilon \right) &\le \delta \, .
%}

\subsection{Performance Measures}\label{sub:perfmeasure}
We are now ready to provide a precise definition of our notion of loss, which would serve as the performance measure of a given algorithm. Given $n\in \NN$, we define the loss of an adaptive algorithm $\Acal$ as:
\als{
L_n(\Acal) := \max_{k\in [K]} L_{k,n}, \quad \hbox{with}\quad  L_{k,n}:=\sum_{x\in \Scal} \hat \pi_{k,n}(x) \|P_k(x,\cdot) - \widehat P_{k,n}(x,\cdot)\|_2^2 \, .
}
The use of the $L_2$-norm in the definition of loss is quite natural in the context of learning and estimation of distributions, as it is directly inspired by the quadratic estimation error used in active bandit allocation (e.g., \cite{carpentier2011upper}).
%where $D(p,q)$ denotes some \emph{contrast measure} or \emph{distance} between distributions $p$ and $q$ defined on the same alphabet.
%redundant? In words, $D(p,q)$ characterizes the discrepancy between $p$ and $q$.
%In this paper, we focus on the following choices for $D$: (i) \emph{Squared distance}: $D_2(p, q) = \|p - q\|_2^2$ and
%\emph{(ii) $\chi^2$ distance}: $D_{\chi^2}(p, q) = \sum_{x\in \Scal} \frac{(p(x) - q(x))^2}{q(x)}$.
%Other candidates for $D$ include, for example, the total variation distance, the Hellinger distance, the Kullback-Leibler (KL) divergence, and the $\chi^2$-distance.
 %The $\chi^2$ distance is also a relevant contrast measures in the statistical estimation of the Markov chains; see, e.g., \cite{billingsley1961statistical}. It also naturally conveys this notion  of quadratic error while being closely related to the KL-divergence.
Given a budget $n$, the loss $L_n(\Acal)$ of an adaptive algorithm $\cA$ is a random variable, due to the evolution of the various chains as well as the possible randomization in the algorithm. Here, we aim at controlling this random quantity in a \emph{high probability} setup as follows: Let $\delta\in (0,1)$. For a given algorithm $\Acal$,  we wish to find $\epsilon:=\epsilon(n,\delta)$ such that
\begin{align}
\label{eq:def_loss_PAC}
\PP\left( L_n(\Acal) \ge \epsilon \right) &\le \delta \, .
\end{align}

\begin{remark}
We remark that the empirical stationary distribution $\hat\pi_{k,t}$ may differ from the stationary distribution associated to the smoothed estimator $\widehat P_{k,t}$ of the transition matrix. Our algorithm and results, however, do not rely on possible relations between $\hat\pi_{k,t}$ and $\widehat P_{k,t}$, though one could have used smoothed estimators for $\pi_k$. The motivation behind using empirical estimate $\hat\pi_{k,t}$ of $\pi_k$ in $L_n$ is that it naturally corresponds to the occupancy of various states according to a given sample path.
\end{remark}

\paragraph{Comparison with other losses.}
We now turn our attention to the comparison between our loss function and some other possible notions. First, we compare ours to the loss function $L'_n(\Acal) = \max_{k} \sum_{x\in\cS}\|P_k(x,\cdot) - \widehat{P}_{k,n}(x,\cdot)\|_2^2$.
%where $D'$ is some proper metric measuring the discrepancy between $P_k$ and $\widehat P_{n,k}$; e.g., $D'(p,q)=\|p- q\|_1$.
Such a notion of loss might look more natural or simpler, since the weights $\hat\pi_{k,n}(x)$ are replaced simply with $1$ (equivalently, uniform weights). However, this means a strategy may incur a high loss for a part of the state space that is rarely visited, even though we have absolutely no control on the chain. For instance, in the extreme case when some states $x$ are reachable with a very small probability, $T_{k,x,n}$ may be arbitrarily small thus resulting in a large loss $L'_n$ for all algorithms, while it makes little sense to penalize an allocation strategy for these ``virtual" states. Weighting the loss according to the empirical frequency $\hat \pi_{k,n}$ of visits avoids such a phenomenon, and is thus more meaningful.

In view of the above discussion, it is also tempting to replace the empirical state distribution $\hat\pi_{k,n}$ with its expectation $\pi_k$, namely to define a \emph{pseudo-loss} function of the form $L''_n(\Acal) = \max_{k} \sum_x\! \pi_k(x) \|P_k(x,\cdot) \!-\! \widehat{P}_{k,n}(x,\cdot)\|_2^2$ (as studied in, e.g., \cite{hao2018learning} in a different setup).
We recall that our aim is to derive performance guarantees on the algorithm's loss that hold with high probability (for $1-\delta$ portions of the sample paths of the algorithm for a given $\delta$). To this end, $L_n$ (which uses $\hat\pi_{k,n}$) is more natural and meaningful than $L_n''$ as $L_n$ penalizes the algorithm's performance by the relative visit counts of various states in a given sample path (through $\hat\pi_{k,n}$), and not by the expected value of these. This matters a lot in the small-budget regime, where $\hat\pi_{k,n}$ could differ significantly from $\pi_k$ --- Otherwise when $n$ is large enough,  $\hat \pi_{k,n}$ becomes well-concentrated around $\pi_k$ with high probability. To clarify further, let us consider the small-budget regime, and some state $x$ where $\pi_k(x)$ is not small. In the case of $L_n$, using $\hat\pi_{k,n}$ we penalize the performance by the mismatch between $\widehat P_{k,n}(x,\cdot)$ and $P_k(x,\cdot)$, weighted proportionally to the number of rounds the algorithm has actually visited $x$. In contrast, in the case of $L_n''$, weighting the mismatch proportionally to $\pi_k(x)$ does not seem reasonable  since in a given sample path, the algorithm might not have visited $x$ enough even though $\pi_k(x)$ is not small. We remark that our results in subsequent sections easily apply to the pseudo-loss $L''_n$, at the expense of an additive second-order term, which might depend on the mixing times.

Finally, we position the high-probability guarantee on $L_n$, in the sense of Eq.~(\ref{eq:def_loss_PAC}), against those holding in expectation. Prior studies on bandit allocation, such as \cite{antos2010active,carpentier2011upper}, whose objectives involve a max operator, consider expected squared distance. The presented analyses in these series of works rely on Wald's second identity as the technical device. This prevents one to extend the approach therein to other distances.
Another peculiarity arising in working with expectations is the order of ``max'' and ``expectation'' operators. While it makes more sense to control  the \emph{expected value of the maximum}, the works cited above look at \emph{maximum of the expected value}, which is more in line with a pseudo-loss definition rather than the loss. All of these difficulties can be avoided by resorting to a high probability setup (in the sense of Eq.~(\ref{eq:def_loss_PAC}).

\paragraph{Further intuition and example.}
We now provide an illustrative example to further clarify some of the above comments.
Let us consider the following two-state Markov chain: $P =
\begin{bmatrix}
1/2 & 1/2 \\
\epsilon & 1-\epsilon
\end{bmatrix}$\, ,
where $\epsilon\in (0,1)$. The stationary distribution of this Markov chain is $\pi = [\tfrac{\epsilon}{2+\epsilon}, \tfrac{2}{2+\epsilon}]$.
Let $s_1$ (resp.~$s_2$) denote the state corresponding to the first (resp.~second) row of the transition matrix.
In view of $\pi$, when $\epsilon\ll 1$, the chain tends to stay in $s_2$ (the lazy state) most of the time:
Out of $n$ observations, one gets on average only $n\pi(s_1) = n\epsilon/(2+\epsilon)$ observations of state $s_1$,
which means, for $\epsilon \ll 1/n$, essentially no observation of state $s_1$.
Hence, no algorithm can estimate the transitions from $s_1$ in such a setup, and  all strategies would suffer a huge loss according to $L_n'$, no matter how samples are allocated to this chain.
Thus, $L_n'$ has limited interest in order to distinguish between good and base sampling strategies.
On the other hand, using $L_n$ enables to better distinguish between allocation strategies, since the weight given to $s_1$ would be essentially $0$ in this case, thus focusing on the good estimation of $s_2$ (and other chains) only.

\subsection{Static Allocation}\label{sec:static}
%\oam{To be modified}
%
%Note that for any chain $k$ and $x$, $T_{k,x,n}$ is a stopping time. Using the fact that observations in a given state are i.i.d. (given observation at state $x$)~and second Wald's inequality, we can prove the following.
%
%We have
%\als{
%\EE\left[\hat\pi_{k,n}(x) \sum_y (P_k - \widehat P_{k,n})(x,y)^2\right] &= \EE\left[\frac{T_{k,x,n}}{T_{k,n}} \sum_y (P_k - \widehat P_{k,n})(x,y)^2\right] \\
%&= \EE\left[\frac{1}{T_{k,n}T_{k,x,n}} \sum_y \EE\left[\left(T_{k,x,n}P_k(x,y) - T_{k,x,y,n}\right)^2\Big|T_{k,x,n}\right]\right] \\
%&= \EE\left[\frac{1}{T_{k,n}} \sum_y P_k(I-P_k)(x,y)\right] \\
%&= \EE\left[\frac{G_k(x)}{T_{k,n}} \right] \\
%}
%Hence,
%\als{
%\EE[L_{k,n}] = \sum_{x}G_k(x) \EE\left[\frac{1}{T_{k,n}}\right]\, .
%}
%
%Now for the static allocation, $T_{k,n}$ does not depend on data and is deterministic. We have
%\als{
%T_{k,n} = \eta_k n
%}
%where

%\begin{definition}
%  An algorithm $\Acal$ is said to be \emph{asymptotically optimal} if for any problem instance, $L_{n}(\Acal)=o(n^{\alpha})$ for all $\alpha>-1$ with high probability.
%\end{definition}

In this subsection, we investigate the optimal loss asymptotically achievable by an oracle policy that is aware of some properties of the chains. To this aim, let us consider a non-adaptive strategy where sampling of various chains is deterministic. Therefore, $T_{k,n}, k=1,\ldots,K$ are not random. The following lemma is a consequence of the central limit theorem:

%In the case of square loss function, for any chain $k$,
%\als{
%	L_{n,k} = \frac{1}{T_{k,n}}\sum_x \sum_y \Big(\sqrt{T_{k,x,n}}(P_k(x,y) - \widehat{P}_{k,n}(x,y)) \Big)^2
%}
%Now when $T_{k,n}$ grows large, this further implies $T_{k,x,n}\to\infty$ as a consequence of  ergodicity. Hence, by the central limit theorem, for each pair $(x,y)$, $\sqrt{T_{k,x,n}}(P_k(x,y) - \widehat{P}_{k,n}(x,y))$ would be distributed according to a centered Gaussian distribution with variance $P_k(I-P_k)(x,y)$. Assuming for the moment that all variances are equal to the same constant $C$, we then deduce that the quantity $\sum_y \Big(\sqrt{T_{k,x,n}}(P_k(x,y) - \widehat{P}_{k,n}(x,y)) \Big)^2/C$ converges (in distribution) to a $\chi^2$ distribution with degree $S-1$. Thus,  we observe that $L_{k,n}$ approaches $C/T_{k,n}$ times the sum of $\chi^2$ distributed random variables, whose mean is $S-1$.
%More generally, $L_{k,n}$ satisfies\footnote{When all $P_k$ are constant $1/S$, $G_k(x)=1-1/S$, and $\sum_x G_k(x) = S-1$.},

\begin{lemma}\label{lem:oracle_loss}
We have for any chain $k$:
%$$
%nL_n(\Acal_{\textrm{oracle}}) \to_{n\to\infty} \Lambda\, .
%$$
%\end{lemma}
$
	T_{k,n}L_{k,n} \to_{T_{k,n}\to\infty} \sum_x G_k(x)\, .
$
\end{lemma}

The proof of this lemma consists in two steps: First, we provide lower and upper bounds on $L_{k,n}$ in terms of the loss $\widetilde L_{k,n}$ incurred by the learner had she used an empirical estimator (corresponding to $\alpha=0$ in (\ref{eq:P_estimator_alpha})). Second, we show that by the central limit theorem,  $T_{k,n}\widetilde L_{k,n} \to_{T_{k,n}\to\infty} \sum_x G_k(x)$.

Now, consider an oracle policy $\Acal_{\mathrm{oracle}}$, who is aware of $\sum_{x\in \Scal} G_k(x)$ for various chains. In view of the above discussion, and taking into account the constraint $\sum_{k\in [K]} T_{k,n}=n$, it would be asymptotically optimal to allocate $T_{k,n}=\eta_k n$ samples to chain $k$, where
$$
\eta_k:=\frac{1}{\Lambda}\sum_{x\in \cS} G_k(x)\, ,\quad \hbox{with} \quad \Lambda := \sum_{k\in [K]} \sum_{x\in \cS} G_k(x)\, .
$$
The corresponding loss would satisfy:
$
nL_n(\Acal_{\textrm{oracle}}) \to_{n\to\infty} \Lambda\, .
$
%\begin{lemma}[{Asymptotically Optimal Loss}]\label{lem:oracle_loss}
%The loss under an oracle algorithm satisfies:
%$$
%nL_n(\Acal_{\textrm{oracle}}) \to_{n\to\infty} \Lambda\, .
%$$
%\end{lemma}
We shall refer to the quantity $\frac{\Lambda}{n}$ as \emph{the asymptotically optimal loss}, which is a \emph{problem-dependent} quantity.
The coefficients $\eta_k, k\in [K]$ characterize the discrepancy between the transition matrices of the various chains, and indicate that an algorithm needs to account for such discrepancy in order to achieve the asymptotically optimal loss.
Having characterized the notion of asymptotically optimal loss, we are now ready to define the notion of \emph{uniformly good algorithm}:

%Next we investigate the optimal loss that can be achieved asymptotically. To this aim, let us consider a non-adaptive strategy where sampling of various chains is not determined by the history. Hence, $T_{k,n},k=1,\cdots,K$ are  not random. Using concentration inequalities, for any chain $k$ one has
%\als{
%\limsup_{n\to\infty} T_{k,n}L_{k,n}\le \sum_{x}G_k(x)
%}
%with high probability.
%Hence, if one knows $\sum_x G_k(x)$ for various chains, and devises an strategy $\Acal_{\textrm{oracle}}$ ensuring
%$$
%T_{k,n} = \frac{\sum_x G_k(x)}{\Lambda} n
%$$
%asymptotically, where $\Lambda := \sum_{k\in [K]} \sum_{x\in \Scal_k} G_k(x)$, then
%\als{
%\limsup_{n\to\infty} n\log(n) L_n(\Acal_{\textrm{oracle}}) \le \Lambda\, .
%}
%
%We shall refer to the quantity $n^{-1}\Lambda$ as the asymptotically optimal loss, which is problem-dependent quantity. We further introduce, for any chain $k$,
%$$
%\eta_k:= \frac{\sum_x G_k(x)}{\Lambda} n.
%$$
%The coefficients $\eta_k, k\in [K]$ characterize the discrepancy between the transition matrix of the various chains, and indicate that an algorithm needs to respect such discrepancy in order to achieve the asymptotically optimal loss.

 %&= \EE\left[\frac{T_{k,x,n}}{T_{k,n}} \sum_y (P_k - \widehat P_{k,n})(x,y)^2\right] \\
%&= \EE\left[\frac{1}{T_{k,n}T_{k,x,n}} \sum_y \EE\left[\left(T_{k,x,n}P_k(x,y) - T_{k,x,y,n}\right)^2\Big|T_{k,x,n}\right]\right] \\
%&= \EE\left[\frac{1}{T_{k,n}} \sum_y P_k(I-P_k)(x,y)\right] \\
%&= \EE\left[\frac{G_k(x)}{T_{k,n}} \right] \\

\begin{definition}[{Uniformly Good Algorithm}]
\label{def:unif_Sq}
  An algorithm $\Acal$ is said to be \emph{uniformly good} if, for any problem instance, it achieves the asymptotically optimal loss when $n$ grows large; that is,
  $\lim_{n\to\infty} nL_n(\Acal)=\Lambda$ for all problem instances.
\end{definition}

%A similar reasoning for the $\chi^2$ criterion leads to the following definition:
%
%\begin{definition}[{Uniformly Good Algorithm, $\chi^2$ Loss}]
%\label{def:unif_xi2}
%  An algorithm $\Acal$ is said to be \emph{uniformly good} w.r.t.~the $\chi^2$ loss if for any problem instance, it achieves the optimal asymptotic loss in the limit with high probability; that is
%  $L_n(\Acal)=S(S-1)n^{-1}$ with high probability as $n$ grows large.
%\end{definition}
 %One simple uniformly good  algorithm consists in sampling chains in a round robin manner. One can show that the loss of this algorithm is $\widetilde\Ocal(n^{-1}K\max_{k} G_k(x))$ with high probability -- see the proof of Theorem \ref{thm:SC_loss_L2}.

\section{The \BAMC\ Algorithm}
\label{sec:algo}
In this section, we introduce an algorithm designed for adaptive bandit allocation of a set of Markov chains.
It is designed based on the \emph{optimistic principle}, as in MAB problems (e.g.,  \cite{lai1985asymptotically,auer2002finite}), and relies on an index function.
More precisely, at each time $t$, the algorithm maintains an index function $b_{k,t+1}$ for each chain $k$, which provides an upper confidence bound (UCB) on the loss incurred by $k$ at $t$; more precisely, with high probability, $b_{k,t+1}\ge L_{k,t}:=\sum_{x\in \Scal} \hat\pi_{k,t}(x)\|P_k(x,\cdot) - \widehat P_{k,t}(x,\cdot)\|_2^2$, where $\widehat{P}_{k,t}$ denotes the smoothed estimate of $P_k$ with some $\alpha>0$ (see Eq.~(\ref{eq:P_estimator_alpha})). Now, by sampling a chain $k_t \in \argmax_{k\in[K]} b_{k,t+1}$ at time $t$, we can balance exploration and exploitation by selecting more the chains with higher estimated
losses or those with higher uncertainty in these estimates.

In order to specify the index function $b_{k,\cdot}$, let us choose $\alpha=\frac{1}{3S}$ (we motivate this choice of $\alpha$ later on), and for each state $x\in \Scal$, define the estimate of Gini coefficient at time $t$ as
$
\widehat G_{k,t}(x):= \sum_{y\in \Scal} \widehat P_{k,t}(x,y) (1 - \widehat P_{k,t}(x,y)).
$
The index $b_{k,t+1}$ is then defined as
\als{
b_{k,t+1} &= \frac{2\beta}{T_{k,t}} \sum_{x\in \Scal} \indic \{T_{k,x,t}>0\}\widehat G_{k,t}(x) +  \frac{6.6\beta^{3/2}}{T_{k,t}}\sum_{x\in \Scal} \frac{T_{k,x,t}^{3/2}}{(T_{k,x,t}+\alpha S)^2}\sum_{y\in \Scal} \sqrt{\widehat{P}_{k,t} (I-\widehat{P}_{k,t})(x,y)} \\
& +
\frac{28\beta^2S}{T_{k,t}} \sum_{x\in \Scal} \frac{\indic \{T_{k,x,t}>0\}}{T_{k,x,t} + \alpha S}
\,  ,
}
where $\beta := \beta(n,\delta) := c\log\Big(\left\lceil\frac{\log(n)}{\log(c)}\right\rceil\frac{6KS^2}{\delta}\Big)$, with $c>1$ being an arbitrary choice. In this paper, we choose $c=1.1$.

We remark that the design of the index $b_{k,\cdot}$ above comes from the application of empirical Bernstein concentration for $\alpha$-smoothed estimators (see Lemma 4 in the supplementary) to the loss function $L_{k,t}$. In other words, Lemma 4 guarantees that with high probability, $b_{k,t+1}\ge L_{k,t}$. Our concentration inequality (Lemma 4) is new, to our knowledge, and could be of independent interest.

Having defined the index function $b_{k,\cdot}$, we are now ready to describe our algorithm, which we call  \BAMC\ (Bandit Allocation for Markov Chains).
\BAMC\ receives as input a confidence parameter $\delta$, a budget $n$, as well as the state space $\Scal$. It initially samples each chain twice (hence, this phase lasts for $2K$ rounds). Then, \BAMC\ simply consists in sampling the chain with the largest index $b_{k,t+1}$ at each round $t$. Finally, it returns, after $n$ pulls, an estimate  $\widehat P_{k,n}$ for each chain $k$.
We provide the pseudo-code of \BAMC\ in Algorithm~\ref{alg:algo_loss_ii}.
Note that \BAMC\ does not require any prior knowledge of the chains (neither the initial distribution nor the mixing time).

%Note that since the performance is measured by the maximum error on each chain, the loss does \emph{not} decrease unless all states in all chains have been visited at least once.

%\sadegh{To be mention that init part is not limiting, as for variance estimation at least two samples are needed.}

\begin{algorithm}[h]
   \caption{\BAMC\ -- Bandit Allocation for Markov Chains}
   \label{alg:algo_loss_ii}
\begin{algorithmic}
   \STATE \textbf{Input:} Confidence parameter $\delta$, budget $n$, state space $\Scal$;
   \STATE \textbf{Initialize:} Sample each chain twice;
      \FOR{$t=2K+1,\ldots,n$}
        \STATE Sample chain $k_t \in\argmax_{k} b_{k,t+1}$;
        \STATE Observe $X_{k,t}$, and update $T_{k,x,t}$ and $T_{k,t}$;
   \ENDFOR
\end{algorithmic}
\normalsize
\end{algorithm}

In order to provide more insights into the design of \BAMC, let us remark that (as shown in Lemma 8 in the supplementary) $b_{k,t+1}$ provides a high-probability UCB on the quantity $\frac{1}{T_{k,t}}\sum_{x} G_k(x)$ as well. Now by sampling the chain $k_t \in \argmax_{k\in[K]} b_{k,t+1}$ at time $t$, in view of discussions in Section \ref{sec:static}, \BAMC\ would try to mimic an oracle algorithm being aware of $\sum_x G_k(x)$ for various chains.

We remark that our concentration inequality in Lemma 4 (of the supplementary) parallels the one presented in Lemma 8.3 in \cite{hsu2015mixing}. In contrast, our concentration lemma makes appear the terms $T_{k,x,t} + \alpha S$ in the denominator, whereas Lemma 8.3 in \cite{hsu2015mixing} makes appear terms $T_{k,x,t}$ in the denominator. This feature plays an important role to deal with situations where some states are not sampled up to time $t$, that is for when $T_{k,x,t}=0$ for some $x$.

%To state the performance guarantee of BA-MC, we introduce the following notation.
%Define $N_{\textrm{init}} := \inf\{t: T_{k,x,t}\ge 2, \forall k, \forall x\}$.

\section{Performance Bounds}\label{sec:bounds}
We are now ready to study the performance bounds on the loss $L_n(\text{\BAMC})$ in both asymptotic and non-asymptotic regimes. We begin with a generic non-asymptotic bound as follows:

\begin{theorem}[\BAMC, Generic Performance]
 \label{thm:loss_squared_generic}
Let $\delta\in (0,1)$. Then, for any budget $n\ge 4K$, with probability at least $1-\delta$,
the loss under $\Acal=\BAMC$ satisfies
\als{
  L_n(\Acal) \le \frac{304KS^2\beta^2}{n} + \widetilde \cO\Big(\frac{K^2S^2}{n^2}\Big)\, .
}
\end{theorem}

The proof of this theorem, provided in Section C in the supplementary, reveals the motivation to choose $\alpha = \frac{1}{3S}$: It verifies that to minimize the dependency of the loss on $S$, on must choose   $\alpha \propto S^{-1}$. In particular, the proof does not rely on the ergodicity assumption:

\begin{remark}
Theorem \ref{thm:loss_squared_generic} is valid even if the Markov chains $P_k,k\in [K]$ are reducible or periodic.
\end{remark}

In the following theorem, we state another non-asymptotic bound on the performance of \BAMC, which refines Theorem \ref{thm:loss_squared_generic}. To present this result, we recall the notation $\Lambda := \sum_{k}\sum_x G_k(x)$, and that for any chain $k$, $\eta_k = \frac{1}{\Lambda}\sum_{x\in \cS} G_k(x)$, $H_k   :=  \sum_{x\in \Scal} \pi_k(x)^{-1}$, and $\underline{\pi}_k := \min_{x\in \cS} \pi_k(x) >0.$
%In the following theorem, we provide a second non-asymptotic bound on the loss, which is valid when the budget $n$ is sufficiently large.

\begin{theorem}%[\BAMC, Large Budget Regime]
 \label{thm:loss_squared_suff_budget}
Let $\delta\in (0,1)$, and assume that $n\ge\text{\ncut}: = K\max_k \Big(\frac{300}{\gamma_{\textsf{ps},k}\underline{\pi}_k}\log\Big(\frac{2K}{\delta}\sqrt{\underline{\pi}_k^{-1}}\Big)\Big)^2.
$
Then, with probability at least $1-2\delta$,
\als{
L_{n}(\Acal) &\le \frac{2\beta \Lambda}{n} + \frac{C_0\beta^{3/2}}{n^{3/2}}  + \widetilde\Ocal(n^{-2})\, ,
}
where
$
C_0:=150K\sqrt{S\Lambda \max_k H_k} + 3\sqrt{S\Lambda}\max_k \frac{H_k}{\eta_k}.
$
\end{theorem}
%The above theorem implies: with high probability,
%$
%n^{3/2}\left(L_n - n^{-1}\beta \Lambda\right)\to_{n\to\infty}
%\widetilde{\Ocal}(S^{1/2}\Lambda^{5/2}\max_k\sqrt{H_k}/\eta_k).
%$

Recalling the asymptotic loss of the oracle algorithm discussed in Section \ref{sec:static} being equal to $\Lambda/n$, in view of the Bernstein concentration, the oracle would incur a loss at most $\frac{2\beta\Lambda}{n}$ for when the budget $n$ is finite. In this regard, we may look at the quantity  $L_n(\cA) - \frac{2\beta\Lambda}{n}$ as the \emph{pseudo-excess loss} of $\cA$ (we refrain from calling this quantity the \emph{excess loss}, as $\frac{2\beta\Lambda}{n}$ is not equal to the high-probability loss of the oracle).
Theorem \ref{thm:loss_squared_suff_budget} implies that when $n$ is greater than the cut-off budget \ncut, the pseudo-excess loss under \BAMC\ vanishes at a rate $\widetilde\cO(n^{-3/2})$. In particular, Theorem \ref{thm:loss_squared_suff_budget} characterizes the constant $C_{0}$ controlling the main term of the pseudo-excess loss: $C_{0}=\cO(K\sqrt{S\Lambda \max_k H_k} + \sqrt{S\Lambda}\max_k \frac{H_k}{\eta_k})$. This further indicates that  the pseudo-excess loss is controlled by the quantity $\frac{H_k}{\eta_k}$, which captures (i) the discrepancy among the $\sum_x G_k(x)$ values of various chains $k$, and (ii)  the discrepancy between various stationary probabilities $\pi_k(x),x\in \Scal$. We emphasize that the dependency of the learning performance (through $C_0$) on $H_k$ is in alignment with the result obtained by \cite{wolfer2019minimax} for the estimation of a single ergodic Markov chain.

%Hence, the loss approaches that of an oracle $\Lambda/n$, up to a multiplicative $\cO(\log(\log(n))$ factor, as well as additive terms vanishing at rate $\cO(n^{-3/2})$. % We also remark that we may extend the bound in Theorem \ref{thm:loss_squared_suff_budget} to  non-reversible chains by replacing the spectral gap $\gamma_k$ with pseudo-spectral gap $\gamma_{\textsf{ps}}$. %{\color{blue}invoking the corresponding concentration inequality.}

The proof of this theorem, provided in Section D in the supplementary, shows that to determine the cut-off budget \ncut, one needs to determine the value of $n$ such that with high probability, for any chain $k$ and state $x$, the term $T_{k,n}(T_{k,x,n}+\alpha S)^{-1}$ approaches $\pi_k(x)^{-1}$, which is further controlled by $\gamma_{\textsf{ps},k}$ (or $\gamma_{k}$ if $k$ is reversible) as well as the minimal stationary distribution $\underline\pi_k$. This in turn allows us to show that, under \BAMC, the number $T_{k,n}$ of samples for any chain $k$ comes close to the quantity $\eta_k n$.
Finally, we remark that the proof of Theorem \ref{thm:loss_squared_suff_budget} also reveals that the result in the theorem is indeed valid for any constant $\alpha>0$.

In the following theorem, we characterize the asymptotic performance of \BAMC:

\begin{theorem}[\BAMC, Asymptotic Regime]
 \label{thm:loss_squared_asymp}
Under $\Acal=$\BAMC, $
\limsup_{n\to\infty} nL_n(\cA) = \Lambda\, .
$
\end{theorem}

The above theorem asserts that, asymptotically, the loss under \BAMC\ matches the asymptotically optimal loss $\Lambda/n$ characterized in Section \ref{sec:static}. We may thus conclude that \BAMC\ is uniformly good (in the sense of Definition~\ref{def:unif_Sq}). The proof of Theorem \ref{thm:loss_squared_asymp} (provided in Section E of the supplementary) proceeds as follows: It divides the estimation problem into two consecutive sub-problems, the one with the budget $n_0=\sqrt{n}$ and the other with the rest $n-\sqrt{n}$ of pulls. We then show  when $n_0 = \sqrt{n}\ge \text{\ncut}$, the number of samples on each chain $k$ at the end of the first sub-problem is lower bounded by $\Omega(n^{1/4})$, and as a consequence,  the index $b_k$  would be accurate enough:
$b_{k,n_0}\in \frac{1}{T_{k,n_0}}(\sum_x G_k(x), \sum_x G_k(x) + \widetilde\cO(n^{-1/8}))$ with high probability. This allows us to relate the allocation under \BAMC\ in the course of the second sub-problem to that of the oracle, and further to show that the difference vanishes as $n\to\infty$.

Below, we provide some further comments about the presented bounds in Theorems \ref{thm:loss_squared_generic}--\ref{thm:loss_squared_asymp}:

\paragraph{Various regimes.} Theorem \ref{thm:loss_squared_generic} provides a non-asymptotic bound on the loss valid for any $n$, while Theorem \ref{thm:loss_squared_asymp} establishes the optimality of \BAMC\ in the asymptotic regime. In view of the inequality $\Lambda \le K(S-1)$, the bound in Theorem \ref{thm:loss_squared_generic} is at least off by a factor of $S$ from the asymptotic loss $\Lambda/n$. %In particular, when $\Lambda$ is small, the bound in Theorem \ref{thm:loss_squared_generic} is too loose.
Theorem \ref{thm:loss_squared_suff_budget} bridges between the two results thereby establishing a third regime, in which the algorithm enjoys the asymptotically optimal loss up to an additive pseudo-excess loss scaling as $\widetilde\cO(n^{-3/2})$.

\paragraph{The effect of mixing.} It is worth emphasizing that the mixing times of the chains \emph{do not} appear explicitly in the bounds, and only control (through the pseudo-spectral gap $\gamma_{\textsf{ps},k}$) the cut-off budget \ncut\ that ensures when the pseudo-excess loss vanishes at a rate $n^{-3/2}$.
This is indeed a strong aspect of our results due to our meaningful definition of loss, which could be attributed to the fact that our loss function employs  empirical estimates $\hat\pi_{k,n}$ in lieu of $\pi_{k}$. Specifically speaking, as argued in \cite{hsu2015mixing}, given the number of samples of various states (akin to using $\hat\pi_{k,t}(x)$ in the loss definition), the  convergence of frequency estimates towards the true values is independent of the mixing time of the chain.
We note that despite the dependence of \ncut\ on the mixing times, \BAMC\ does not need to estimate them as when $n\le \text{\ncut}$, it still enjoys the loss guarantee of Theorem \ref{thm:loss_squared_generic}. We also mention that to define an index function for the loss function $\max_{k} \sum_x\!
\pi_k(x) \|P_k(x,\cdot) \!-\! \widehat{P}_{k,n}(x,\cdot)\|_2^2$, one may have to derive confidence bounds on the mixing time and/or stationary distribution $\pi_k$ as well.

\paragraph{More on the pseudo-excess loss.}
We stress that the notion of the pseudo-excess loss bears some similarity with the definition of regret for active bandit learning of distributions as introduced in   \cite{antos2010active,carpentier2011upper} (see Section \ref{sec:intro}). In the latter case, the regret typically decays as $n^{-3/2}$ similarly to the pseudo-excess loss in our case.
An interesting question is whether the decay rate of the pseudo-excess loss, as a function of $n$,  can be improved. And more importantly, if a (problem-dependent) lower bound on the pseudo-excess loss can be established. These questions are open even for the simpler case of active learning of distributions in the i.i.d.~setup; see, e.g., \cite{carpentier2014minimax,carpentier2015adaptive,carpentier2011upper}. We plan to address these as a future work.

\section{Conclusion}
In this paper, we addressed the problem of active bandit allocation in the case of discrete and ergodic Markov chains.
We considered a notion of loss function appropriately extending the loss function for learning distributions to the case of Markov chains. We further characterized the notion of a ``uniformly good algorithm'' under the considered loss function. We presented an algorithm for learning Markov chains, which we called \BAMC. Our algorithm is simple to implement and does not require any prior knowledge of the Markov chains. We provided non-asymptotic PAC-type bounds on the loss incurred by \BAMC, and showed that asymptotically, it incurs an optimal loss. We further discussed that the (pseudo-excess) loss incurred by \BAMC\ in our bounds does not deteriorate with mixing times of the chains. %Finally, we provided a simple heuristic to tailor our algorithm for the case of restless chains.
As a future work, we plan to derive a (problem-dependent) lower bound on the pseudo-excess loss.
Another interesting, yet very challenging, future direction is to devise adaptive learning algorithms for restless Markov chains, where the state of various chains evolve at each round independently of the learner's decision.

\section*{Acknowledgements}
This work has been supported by CPER Nord-Pas-de-Calais/FEDER DATA Advanced data science and technologies 2015-2020, the French Ministry of Higher Education and Research, Inria, and the French Agence Nationale de la Recherche (ANR), under grant ANR-16-CE40-0002 (project BADASS).

\bibliography{Bandit_RL_bib_NeurIPS}
\bibliographystyle{unsrt}

\newpage

\appendix

\section{Concentration Inequalities}\label{sec:concentration}

\begin{lemma}[{\cite[Lemma~2.4]{maillard2019mathematics}}]
\label{lem:time_peeling}
Let $Z = (Z_t)_{t\in \NN}$ be a sequence of random variables generated by a predictable process, and $\cF=(\cF_t)_{t}$ be its natural filtration. Let $\phi:\RR\to \RR_+$ be a convex upper-envelope of the cumulant generating function of the conditional distributions with $\phi(0)=0$, and let $\phi_\star$ denote its Legendre-Fenchel transform, that is:
\als{
\forall \lambda\in \cD, \forall t, \qquad &\log\EE\left[\exp\big(\lambda Z_t\big)|\cF_{t-1}\right] \le \phi(\lambda)\, , \\
\forall x\in \RR, \qquad &\phi_\star(x) = \sup_{\lambda\in \RR} (\lambda x - \phi(\lambda))\, ,
}
where $\cD = \{\lambda\in \RR: \forall t, \log\EE\left[\exp(\lambda Z_t)|\cF_{t-1}\right] \le \phi(\lambda) <\infty\}$. Assume that $\cD$ contains an open neighborhood of $0$.
Let  $\phi_{\star,+}^{-1}:\Real\to\Real_+$ (resp.~$\phi_{\star,-}^{-1}$) be its reverse map on $\Real_+$ (resp.~$\Real_-$), that is
\als{
\phi^{-1}_{\star,-}(z):=\sup\{x\le 0: \phi_\star(x)>z\} \quad \hbox{and} \quad \phi^{-1}_{\star,+}(z):=\inf\{x\ge 0: \phi_\star(x)>z\}\, .
}
Let $N_n$ be an integer-valued random variable that is $\cF$-measurable and almost surely bounded by $n$. Then, for all $c\in (1,n]$ and $\delta\in (0,1)$,
\als{
\PP\bigg[\frac{1}{N_n}\sum_{t=1}^{N_n}Z_t \geq  \phi_{\star,+}^{-1}\left(\frac{c}{N_n}\log\Big(\left\lceil\frac{\log(n)}{\log(c)}\right\rceil\frac{1}{\delta}\Big)\right) \bigg] &\leq \delta\, ,\\
\PP\bigg[\frac{1}{N_n}\sum_{t=1}^{N_n}Z_t \leq  \phi_{\star,-}^{-1}\left(\frac{c}{N_n}\log\Big(\left\lceil\frac{\log(n)}{\log(c)}\right\rceil\frac{1}{\delta}\Big)\right) \bigg] &\leq \delta\,  .
}
Moreover, if $N$ is a possibly unbounded $\NN$-valued random variable that is $\cF$-measurable, then for all $c>1$ and $\delta\in (0,1)$,
\als{
\PP\bigg[\frac{1}{N}\sum_{t=1}^{N}Z_t \geq  \phi_{\star,+}^{-1}\left(\frac{c}{N}\log\bigg[\frac{\log(N)\log(cN)}{\delta\log^2(c)}\bigg]\right) \bigg] &\leq \delta\, ,\\
\PP\bigg[\frac{1}{N}\sum_{t=1}^{N}Z_t \leq  \phi_{\star,-}^{-1}\left(\frac{c}{N}\log\bigg[\frac{\log(N)\log(cN)}{\delta\log^2(c)}\bigg]\right) \bigg] &\leq \delta\,  .
}
\end{lemma}

We provide an immediate consequence of this lemma for the case of \emph{sub-Gamma} random variables:

\begin{corollary}
\label{corr:Bernstein_peeling}
  Let $Z = (Z_t)_{t\in \NN}$ be a sequence of random  variables generated by a predictable process, and $\cF=(\cF_t)_{t}$ be its natural filtration. Assume for all $t\in \NN$, $|Z_t|\le b$ and $\EE[Z_{s}^2|\cF_{s-1}] \le v$ for some positive numbers $v$ and $b$.
  Let $N_n$ be an integer-valued random variable that is $\cF$-measurable and almost surely bounded by $n$. Then, for all $c\in (1,n]$ and $\delta\in (0,1)$,
  \als{
\PP\bigg[\frac{1}{N_n}\sum_{t=1}^{N_n}Z_t
\geq \sqrt{\frac{2\zeta(n,\delta) v}{N_n}} +  \frac{\zeta(n,\delta)b}{3N_n} \bigg] &\leq \delta\, , \\
\PP\bigg[\frac{1}{N_n}\sum_{t=1}^{N_n}Z_t \leq -\sqrt{\frac{2\zeta(n,\delta) v}{N_n}} -  \frac{\zeta(n,\delta)b}{3N_n} \bigg]  &\leq \delta\,  ,
}
where $\zeta(n,\delta):= c\log\Big(\left\lceil\frac{\log(n)}{\log(c)}\right\rceil\frac{1}{\delta}\Big)$, with $c>1$ being an arbitrary parameter.
\end{corollary}

\bp
The proof follows by an application of Lemma \ref{lem:time_peeling} for sub-Gamma random variables with parameters $(v,b)$. Note that sub-Gamma random variables satisfy
$
\phi(\lambda) \leq \frac{\lambda^2 v}{2(1-b\lambda)}$, for all $\lambda\in(0,1/b)$,
so that
\als{
    \phi_{\star,+}^{-1}(z) = \sqrt{2v z} + bz   \quad \hbox{and} \quad
    \phi_{\star,-}^{-1}(z) = -\sqrt{2v z} - bz    \, .
}
Plugging these into the first statements of Lemma \ref{lem:time_peeling} completes the proof.
\ep

As a consequence of this corollary, we present the following lemma:

\begin{lemma}[{Bernstein-Markov Concentration}]
\label{lem:time_peeling_Pxy_bernstein}
  Let $(X_t)_{1\le t\le n}$ be generated from an ergodic Markov chain defined on a finite state-space $\Scal$ with transition matrix $P$. Consider the smoothed estimator $\widehat P_{n}$ of $P$ defined as follows: For all $(x,y)\in \cS^2$,
\als{
\widehat P_{n}(x,y):=\frac{\alpha + \sum_{t=2}^n \bI\{X_{t-1}=x, X_{t}=y\}}{\alpha S + T_{x,n}}\, ,
}
with $\alpha >0$. Then, for any $\delta\in (0,1)$, it holds that with probability at least $1-\delta$, for all $(x,y)\in \Scal^2$,
\als{
|\widehat P_n(x,y)- P(x,y)| \le \sqrt{\left(\frac{T_{x,n}}{T_{x,n} + \alpha S}\right)\frac{2P(I- P)(x,y)\zeta(n,\delta)}{T_{x,n}+\alpha S}} + \frac{\tfrac{1}{3}\zeta(n,\delta) + \alpha|1-SP(x,y)|}{T_{x,n} + \alpha S} \, ,
}
where $\zeta(n,\delta):= c\log\Big(\left\lceil\frac{\log(n)}{\log(c)}\right\rceil\frac{2S^2}{\delta}\Big)$, with  $c>1$ being an arbitrary parameter.
\end{lemma}

\bp
The proof uses similar steps as in the one of Lemma 8.3 in \cite{hsu2015mixing}.
Consider a pair $(x,y)\in \cS^2$. We have
\als{
\widehat P_n(x,y)- P(x,y) &= \frac{\alpha + \sum_{t=2}^n \bI\{X_{t-1}=x, X_{t}=y\}}{\alpha S + T_{x,n}} - P(x,y) \\
&= \frac{T_{x,n}}{T_{x,n}+\alpha S}Y_n + \frac{\alpha(1-SP(x,y))}{T_{x,n}+\alpha S}\, ,
}
where $Y_n:=\frac{1}{T_{x,n}}\left(\sum_{t=2}^n \bI\{X_{t-1}=x, X_{t}=y\} - T_{x,n}P(x,y)\right)$. Hence,
\begin{align}
\label{eq:Yn_lemma}
|\widehat P_n(x,y)- P(x,y)| &\le \frac{T_{x,n}}{T_{x,n}+\alpha S}|Y_n| + \frac{\alpha|1-SP(x,y)|}{T_{x,n}+\alpha S}\, .
\end{align}
To control $Y_n$, we define the sequence $(Z_t)_{1\le t\le n}$, with $Z_1:=0$, and
\als{
Z_t := \indic \{X_{t-1}=x\}(\indic \{X_t = y\} - P(x,y)), \quad \forall t\ge 2.
}
Note that for all $t$, $Z_t\in [-P(x,y), 1-P(x,y)]$ almost surely. Moreover, denoting by $(\cF_t)_t$ the filtration generated by $(X_t)_{1\le t\le n}$, we observe that $(Z_t)_{1\le t\le n}$ is $\cF_{t-1}$-measurable and $\EE[Z_t|\cF_{t-1}] = 0$. Hence, it is a martingale difference sequence with respect to $(\cF_t)_t$, and it satisfies $Z_t\in [-P(x,y), 1-P(x,y)]$ for all $t$, and
$$
\EE[Z_t^2|\cF_{t-1}] = P(x,y)(1-P(x,y)) \indic \{X_{t-1}=x\}\, ,\quad \forall t\ge 2.
$$
Applying Corollary \ref{corr:Bernstein_peeling} yields
\als{
|Y_n| \le \sqrt{\frac{2P(I- P)(x,y)\zeta(n,\delta)}{T_{x,n}}} + \frac{\zeta(n,\delta)}{3T_{x,n}} \, ,
}
with probability at least $1-\delta$. Plugging the above bound into (\ref{eq:Yn_lemma}) gives the announced result.
\ep

\begin{lemma}[{Empirical Bernstein-Markov Concentration}]
\label{lem:diff_Pxy_concent_empirical}
Let $(X_t)_{1\le t\le n}$ be generated from an ergodic Markov chain defined on $\Scal$ with transition matrix $P$. Consider the smoothed estimator $\widehat P_{n}$ of $P$ as defined in Lemma \ref{lem:time_peeling_Pxy_bernstein}. Then, with probability at least $1-\delta$, for all $(x,y)\in \Scal^2$,
\als{
|\widehat P_n(x,y)- P(x,y)| \le \left(\frac{2 T_{x,n}\widehat{P}_{n} (I-\widehat{P}_{n})(x,y)\zeta}{(T_{x,n} + \alpha S)^2} +
c_1\frac{\sqrt{T_{x,n}\widehat{P}_{n} (I-\widehat{P}_{n})(x,y)}}{(T_{x,n}+\alpha S)^2}
+ \frac{c_2}{(T_{x,n} + \alpha S)^2}\right)^{1/2}
\, ,
}
where $\zeta:=\zeta(n,\delta):= c\log\Big(\left\lceil\frac{\log(n)}{\log(c)}\right\rceil\frac{2S^2}{\delta}\Big)$, where $c>1$ is an arbitrary parameter, $\zeta':= \tfrac{1}{3}\zeta + \alpha (S-1)$, and
\als{
c_1 = \sqrt{8\zeta}(2\zeta + \zeta')\, \quad\hbox{and}\quad c_2 := \zeta'^2 + 4\zeta(4\zeta + \zeta' + 2\sqrt{\zeta\zeta'}) + \zeta'\sqrt{8\zeta}(5.3\sqrt{\zeta} + \sqrt{2\zeta'}) \, .
}
\end{lemma}

\bp
Fix a pair $(x,y)\in \cS^2$. Recall from Lemma \ref{lem:time_peeling_Pxy_bernstein} that with probability at least $1-\delta$,
\als{
|\widehat P_n(x,y)- P(x,y)|
&\le
\sqrt{\frac{2\zeta T_{x,n}P(I- P)(x,y)}{(T_{x,n}+\alpha S)^2}} + \frac{\zeta'}{T_{x,n} + \alpha S}
\, ,
}
%where {\color{red}$\beta:=...$}, and $\beta':= \tfrac{1}{3}\beta + 2\alpha S$.
so that
\begin{align}
\label{eq:diff_P_xy_1}
(\widehat P_n(x,y)- P(x,y))^2
&\le
\frac{2\zeta T_{x,n}P(I- P)(x,y)}{(T_{x,n}+\alpha S)^2} + \frac{\zeta'^2}{(T_{x,n} + \alpha S)^2} + \sqrt{\frac{8\zeta T_{x,n}P(I- P)(x,y)}{(T_{x,n}+\alpha S)^2}}\frac{\zeta '}{T_{x,n} + \alpha S} \, .
\end{align}

Next we derive an upper bound on $P(I-P)(x,y)$. By Taylor's expansion, we have
\als{
P(I-P)(x,y) &= \widehat{P}_{n} (I-\widehat{P}_{n})(x,y) + (I - 2\widehat{P}_{n})(P - \widehat{P}_{n})(x,y) -(P - \widehat{P}_{n})(x,y)^2 \\
&= \widehat{P}_{n} (I-\widehat{P}_{n})(x,y) + (I - \widehat{P}_{n} - P)(P - \widehat{P}_{n})(x,y) \\
&\le
\widehat{P}_{n} (I-\widehat{P}_{n})(x,y) + |(I - \widehat{P}_{n} - P)(x,y)|\left(\sqrt{\frac{2\zeta  T_{x,n}P(I- P)(x,y)}{(T_{x,n}+\alpha S)^2}} + \frac{\zeta'}{T_{x,n} + \alpha S}\right) \sk
&\le
 \widehat{P}_{n} (I-\widehat{P}_{n})(x,y) + \sqrt{\frac{8\zeta  T_{x,n}P(I- P)(x,y)}{(T_{x,n}+\alpha S)^2}} + \frac{2\zeta '}{T_{x,n} + \alpha S}
\, .
}
Using the fact that $a\le b\sqrt{a}+c$ implies $a\le b^2 + b\sqrt{c} + c$ for nonnegative numbers $a,b,c$, we get
\begin{align}
P(I-P)(x,y)
&\le
\widehat{P}_{n} (I-\widehat{P}_{n})(x,y) + \frac{2\zeta'}{T_{x,n} + \alpha S} +
 \sqrt{\frac{8\zeta  T_{x,n}}{(T_{x,n}+\alpha S)^2}\left(\widehat{P}_{n} (I-\widehat{P}_{n})(x,y) + \frac{2\zeta '}{T_{x,n}+\alpha S}\right)}\sk&
  + \frac{8\zeta  T_{x,n}}{(T_{x,n}+\alpha S)^2} \sk
%&\le
%\widehat{P}_{n} (I-\widehat{P}_{n})(x,y) + \frac{\tilde \zeta }{T_{x,n} + \alpha S}
%+ 2\sqrt{\left(\frac{T_{x,n}}{T_{x,n}+\alpha S}\right)\frac{\zeta \widehat{P}_{n} (I-\widehat{P}_{n})(x,y)}{T_{x,n}+\alpha S}} + \frac{2\tilde \zeta }{T_{x,n}+\alpha S} \\
\label{eq:P_var_1}
&\le
\widehat{P}_{n} (I-\widehat{P}_{n})(x,y) + \sqrt{\frac{8\zeta  T_{x,n}}{(T_{x,n}+\alpha S)^2} \widehat{P}_{n} (I-\widehat{P}_{n})(x,y)} + \frac{8\zeta  + 2\zeta ' + 4\sqrt{\zeta \zeta'}}{T_{x,n}+\alpha S} \, ,
\end{align}
where we have used $\sqrt{a+b}\le \sqrt{a} + \sqrt{b}$ valid for all $a,b\ge 0$. Taking square-root from both sides and using the fact $\sqrt{a+b}\le \sqrt{a}+ \frac{b}{2\sqrt{a}}$ valid for all $a,b>0$ give
\begin{align}
\sqrt{P(I-P)(x,y)}
&\le
\sqrt{\widehat{P}_{n}(I- \widehat{P}_{n})(x,y)} + \frac{1}{\sqrt{T_{x,n}+\alpha S}}\left(\sqrt{2\zeta } + \sqrt{8\zeta  + 2\zeta' + 4\sqrt{\zeta \zeta'}}\right) \sk
\label{eq:P_var_sqrt}
&\le
\sqrt{\widehat{P}_{n}(I- \widehat{P}_{n})(x,y)} + \frac{5.3\sqrt{\zeta } + \sqrt{2\zeta '}}{\sqrt{T_{x,n}+\alpha S}}
\, ,
\end{align}
where we have used
$$
\sqrt{2\zeta} + \sqrt{8\zeta  + 2\zeta' + 4\sqrt{\zeta \zeta'}} \le \sqrt{2\zeta} + \sqrt{6\zeta  + 2(\sqrt{\zeta} + \sqrt{\zeta'})^2} \le 5.3\sqrt{\zeta} + \sqrt{2\zeta'}\, .
$$
Plugging (\ref{eq:P_var_1}) and (\ref{eq:P_var_sqrt}) into (\ref{eq:diff_P_xy_1}), we obtain
\als{
(\widehat P_n(x&,y)- P(x,y))^2 \\
&\le
\frac{2\zeta T_{x,n}}{(T_{x,n} + \alpha S)^2}
\left(
\widehat{P}_{n} (I-\widehat{P}_{n})(x,y) + \sqrt{\frac{8\zeta T_{x,n}}{(T_{x,n}+\alpha S)^2} \widehat{P}_{n} (I-\widehat{P}_{n})(x,y)}
 + \frac{8\zeta + 2\zeta' + 4\sqrt{\zeta\zeta'}}{T_{x,n}+\alpha S}
\right) \\
& + \frac{\zeta'}{T_{x,n}+\alpha S}\sqrt{\frac{8\zeta T_{x,n}}{(T_{x,n} + \alpha S)^2}}\left(\sqrt{\widehat{P}_{n}(I- \widehat{P}_{n})(x,y)} + \frac{5.3\sqrt{\zeta} + \sqrt{2\zeta'}}{\sqrt{T_{x,n} + \alpha S}}\right) + \frac{\zeta'^2}{(T_{x,n} + \alpha S)^2} \\
&\le
\frac{2\zeta T_{x,n}\widehat{P}_{n} (I-\widehat{P}_{n})(x,y)}{(T_{x,n} + \alpha S)^2} +  c_1\frac{\sqrt{T_{x,n}\widehat{P}_{n} (I-\widehat{P}_{n})(x,y)}}{(T_{x,n}+\alpha S)^2}  + \frac{c_2}{(T_{x,n} + \alpha S)^2}
\, ,
}
with
\als{
c_1 &:= \sqrt{8\zeta}(2\zeta + \zeta')\, \quad\hbox{and}\quad c_2 := \zeta'^2 + 4\zeta(4\zeta + \zeta' + 2\sqrt{\zeta\zeta'}) + \zeta'\sqrt{8\zeta}(5.3\sqrt{\zeta} + \sqrt{2\zeta'})\, ,
}
which after taking the square-root from both sides yields the announced result.
\ep

Next we recall a result for the convergence of empirical stationary distributions in a Markov chain to its stationary distribution:

\begin{lemma}[{\cite{paulin2015concentration}}]
\label{lem:stat_concent_Hsu}
Let $(X_t)_{1\le t\le n}$ be an ergodic and reversible Markov chain defined on $\Scal$ with stationary distribution $\pi$ and spectral gap $\gamma$. Let $\hat\pi_n$ denote the corresponding empirical stationary distribution of the Markov chain. For any $\delta\in (0, 1)$, with probability at least $1-\delta$,
\als{
|\hat\pi_n(x) - \pi(x)| \le \sqrt{\frac{8\pi(x)(1-\pi(x))}{\gamma n}\log\left(\frac{1}{\delta}\sqrt{\frac{2}{\min_x \pi(x)}}\right)} + \frac{20}{\gamma n}\log\left(\frac{1}{\delta}\sqrt{\frac{2}{\min_x \pi(x)}}\right)\, , \qquad \forall x\in \Scal.
}
\end{lemma}

\section{Technical Lemmas}

Before providing the proofs of the main theorems, we provide some technical lemmas. We begin with the following definition:

\begin{definition}[{Definition of the Event $C$}]
\label{def:C}
Let $n\ge 1$ and $\delta>0$. For any $(x,y)\in \Scal^2$ and $k\in [K]$ define
\als{
C_{x,y,k}(n,\delta) :=\left\{\forall t\le n: |(\widehat P_{k,t} - P_k)(x,y)| \le \sqrt{\frac{2T_{k,x,t} P_k(I-P_k)(x,y) \beta(n,\delta)}{(T_{k,x,t}+\alpha S)^2}} + \frac{\beta(n,\delta)}{3(T_{k,x,t} + \alpha S)} \right\}\, ,
}
where $\beta(n,\delta):= c\log\Big(\left\lceil\frac{\log(n)}{\log(c)}\right\rceil\frac{6KS^2}{\delta}\Big)$.
Define
$$
C := C(n,\delta) := \cap_{k\in [K]} \cap_{x,y\in \Scal} C_{x,y,k}(n,\delta) \, .
$$
\end{definition}

\begin{lemma}
\label{lem:P_event_C}
For any $n\ge 1$ and $\delta\in (0,1)$, it holds that $\PP(C(n,\delta)) \ge 1-\delta$.
\end{lemma}

\bp
Let $n\ge 1$ and $\delta>0$. Define $\zeta(n,\delta) = c\log\Big(\left\lceil\frac{\log(n)}{\log(c)}\right\rceil\frac{2KS^2}{\delta}\Big)$, and note that  $\beta(n,\delta)=\zeta(n,\delta) + c\log(3)$. Applying Lemma \ref{lem:time_peeling_Pxy_bernstein}, we obtain
\als{
|(\widehat P_{k,t} - P_k)(x,y)| \le \sqrt{\frac{2 T_{k,x,t}P_k(I-P_k)(x,y) \zeta(n,\delta)}{(T_{k,x,t}+\alpha S)^2}} + \frac{\tfrac{1}{3}\zeta(n,\delta)+\alpha (S-1)}{T_{k,t,x}+\alpha S} \, ,
}
for all $(x,y)$, and uniformly for all $t\le n$, with probability at least $1-\frac{\delta}{K}$. With the choice of $\alpha = \frac{1}{3S}$, and noting that $\beta(n,\delta)\ge\zeta(n,\delta)$ and
$$
\frac{\zeta(n,\delta)}{3} + \frac{S-1}{3S} \le \frac{\beta(n,\delta)}{3},
$$
we obtain $\PP(\cap_{x,y\in \cS} C_{x,y,k}(n,\delta)) \ge 1-\delta/K$ for all $k$.
Finally, using a union bound gives $\PP(C(n,\delta)) \ge 1-\delta$.
\ep

In the following lemma we provide an upper bound on the loss $L_{k,n}$, which is valid for all $n$.

\begin{lemma}[{Upper Bound on the Loss}]
\label{lem:L2_UB_any_budget}
Assume that the event $C$ holds. Then, for any budget $n>1$ and chain $k$,
\als{
L_{k,n} \le \frac{2\beta}{T_{k,n}}\sum_{x}  G_k(x) \indic \{T_{k,x,n}>0\}
+ \frac{2\sqrt{2}}{3}\frac{\beta^{3/2}\sqrt{S}}{T_{k,n}}\sum_x \frac{T_{k,x,n}^{3/2}\sqrt{G_k(x)}}{(T_{k,x,n}+\alpha S)^2} +
\frac{S\beta^2}{9T_{k,n}}\sum_x \frac{T_{k,x,n}}{(T_{k,x,n} + \alpha S)^2} \,.
}
\end{lemma}

\bp
Let $n>1$ and consider a chain $k$. To simplify the notation, we omit the dependence of various quantities on $k$ (hence $T_{x,n}:=T_{k,x,n}$, $T_{n}:=T_{k,n}$, and so on). On the event $C$, we have
\als{
|\widehat P_n(x,y)- P(x,y)| &\le \sqrt{\left(\frac{T_{x,n}}{T_{x,n} + \alpha S}\right)\frac{2 P(I- P)(x,y)\beta}{T_{x,n}+\alpha S}} + \frac{\beta}{3(T_{x,n} + \alpha S)}
\, ,
}
so that
\als{
(\widehat P_n(x,y)- P(x,y))^2
&\le
\frac{2\beta P(I- P)(x,y)}{T_{x,n}+\alpha S} + \frac{\tfrac{1}{9}\beta^2}{(T_{x,n} + \alpha S)^2} + \frac{\tfrac{2\sqrt{2}}{3}\beta^{3/2}}{(T_{x,n} + \alpha S)^2}\sqrt{T_{x,n}P(I- P)(x,y)}\, .
}
Hence, we obtain the announced upper bound on the loss:
\als{
L_{n} &= \sum_x \hat\pi_{n}(x)\sum_y (\widehat P_n(x,y)- P(x,y))^2 \\
&\le
\frac{2\beta}{T_n}\sum_{x} \frac{T_{x,n}G(x)}{T_{x,n}+\alpha S}
+ \frac{2\sqrt{2}}{3}\frac{\beta^{3/2}}{T_n}\sum_x \frac{T_{x,n}^{3/2}}{(T_{x,n}+\alpha S)^2} \sum_y \sqrt{P(I- P)(x,y)} +
\frac{S\beta^2}{9T_n}\sum_x \frac{T_{x,n}}{(T_{x,n} + \alpha S)^2} \\
&\le
\frac{2\beta}{T_n}\sum_{x} G(x)\indic \{T_{x,n}>0\}
+ \frac{2\sqrt{2}}{3}\frac{\beta^{3/2}\sqrt{S}}{T_n}\sum_x \frac{T_{x,n}^{3/2} \sqrt{G(x)}}{(T_{x,n}+\alpha S)^2} +
\frac{S\beta^2}{9T_n}\sum_x \frac{ T_{x,n}}{(T_{x,n} + \alpha S)^2} \, ,
}
where the last step follows from the Cauchy-Schwarz inequality.
\ep

The following lemma presents bounds on the index $b_{k,\cdot}$ on the event $C$ (defined in Definition \ref{def:C}):

\begin{lemma}[{Bounds on the Index}]
\label{lem:index_Sq_UB_LB_any_budget}
Consider a chain $k$, and assume that the event $C$ holds. Then, for any time $t$,
\als{
b_{k,t+1} &\le \frac{2\beta}{T_{k,t}}\sum_x G_k(x)\indic \{T_{k,x,t}>0\}
+ \frac{13\beta^{3/2}\sqrt{S}}{T_{k,t}}\sum_x \sqrt{\frac{G_k(x)\indic \{T_{k,x,t}>0\}}{T_{k,x,t} + \alpha S}}
+ \frac{39\beta^2 S}{T_{k,t}}\sum_x \frac{\indic \{T_{k,x,t}>0\}}{T_{k,x,t} + \alpha S}\, ,\\
b_{k,t+1} &\ge \frac{2\beta}{T_{k,t}}\sum_x G_k(x)\indic \{T_{k,x,t}>0\} \, .
}
%with
%\als{
%c_3 &= c_1 + 2\sqrt{8}\beta^{3/2}\, \quad \hbox{and} \quad c_4 = c_2 + \sqrt{2} c_1(\sqrt{\beta} + \sqrt{\beta'}) + 4\beta\beta'\, .
%}
\end{lemma}

\bp
Fix a chain $k$ and time $t$. To ease notation, let us omit the dependence of various quantities on $k$ throughout.
We first recall the definition of the index $b_{t+1}$:
\als{
b_{t+1} = \frac{2\beta}{T_t} \sum_x \widehat G_t(x)\indic \{T_{x,t}>0\} +  \frac{c_1}{T_t}\sum_x \frac{T_{x,t}^{3/2}}{(T_{x,t}+\alpha S)^2}\sum_y \sqrt{\widehat{P}_{t} (I-\widehat{P}_{t})(x,y)} +
\frac{c_2S}{T_t} \sum_{x} \frac{\indic \{T_{x,t}>0\}}{T_{x,t} + \alpha S}
\, ,
}
where $c_1=6.6\beta^{3/2}$ and $c_2=28\beta^2$.

To derive an upper bound on $b_{t}$, we first find an upper bound on $\widehat P_{t}(I-\widehat P_{t})(x,y)$ as follows. First, using Taylor's expansion, we have
\begin{align}
\widehat P_t(I-\widehat P_t)(x,y) &= P(I-P)(x,y) + (I - 2P)(\widehat{P}_{t} - P)(x,y) -( \widehat{P}_{t} - P)(x,y)^2 \sk
&= P(I-P)(x,y) + (I - P - \widehat{P}_{t})(\widehat{P}_{t}-P)(x,y) \sk
\label{eq:UB_var_P}
&\le
P(I-P)(x,y) + \sqrt{\frac{8\beta P(I- P)(x,y)}{T_{x,t} + \alpha S}} + \frac{2\beta}{3(T_{x,t} + \alpha S)} \\
&\le \left(\sqrt{P(I-P)(x,y)} + \sqrt{\frac{2\beta}{T_{x,t}+\alpha S}}\right)^2\, ,\nonumber
\end{align}
where (\ref{eq:UB_var_P}) follows from the definition of $C$. Hence,
\begin{align}
\label{eq:UB_sqrt_var_P}
\sqrt{\widehat P_{t}(I-\widehat P_t)(x,y)}
&\le \sqrt{P(I- P)(x,y)} + \sqrt{\frac{2\beta}{T_{x,t} + \alpha S}} \, .
\end{align}
% where we used inequalities $\sqrt{a+b}\le \sqrt{a} + \sqrt{b}$ and $\sqrt{a+b}\le \sqrt{a}+ \frac{b}{2\sqrt{a}}$ valid for all $a,b>0$.
Using (\ref{eq:UB_var_P}) and (\ref{eq:UB_sqrt_var_P}), we obtain the following upper bound on $b_t$, on the event $C$:
\als{
b_{t+1} &\le \frac{2\beta}{T_{t}}\sum_x  \sum_y \left(P(I-P)(x,y) + \sqrt{\frac{8\beta P(I- P)(x,y)}{T_{x,t} + \alpha S}} + \frac{2\beta}{3(T_{x,t} + \alpha S)}\right) \indic \{T_{x,t}>0\}\\
&+  \frac{6.6\beta^{3/2}}{T_{t}}\sum_x \frac{T_{x,t}^{3/2}}{(T_{x,t}+\alpha S)^2} \sum_y \left(\sqrt{P(I- P)(x,y)} + \frac{\sqrt{2\beta}}{\sqrt{T_{x,t} + \alpha S}}\right)
+ \frac{28\beta^2S}{T_{t}}\sum_x \frac{\indic \{T_{x,t}>0\}}{T_{x,t} + \alpha S}\\
&\le
\frac{2\beta}{T_{t}}\sum_x G(x)\indic \{T_{x,t}>0\}
+ \frac{13\beta^{3/2}\sqrt{S}}{T_{t}}\sum_x \sqrt{\frac{G(x)\indic \{T_{x,t}>0\}}{T_{x,t} + \alpha S}}
+ \frac{39\beta^2 S}{T_{t}}\sum_x \frac{\indic \{T_{x,t}>0\}}{T_{x,t} + \alpha S}\, .
}

To prove the lower bound on the index, we recall from the proof of Lemma \ref{lem:diff_Pxy_concent_empirical} (see (\ref{eq:P_var_1}) with the choices $\zeta = \beta$ and $\zeta'=\frac{\beta}{3}$) that
\begin{align*}
\widehat{P}_{t} (I-\widehat{P}_{t})(x,y)
&\ge
P(I-P)(x,y) - \sqrt{\frac{8\beta  T_{x,t}}{(T_{x,t}+\alpha S)^2} \widehat{P}_{t} (I-\widehat{P}_{t})(x,y)} - \frac{12}{T_{x,t}+\alpha S} \, .
\end{align*}
Putting this together with the definition of $b_{t+1}$ leads to $b_{t+1} \ge \frac{2\beta}{T_{t}}\sum_x G(x)\indic \{T_{x,t}>0\}$, and thus completes the proof.
\ep

\section{Proof of Theorem \ref{thm:loss_squared_generic}}\label{sec:proof_generic}

Consider a chain $k$ %(and so let us omit the dependence on $k$ unless stated otherwise).
and assume that the event $C$ (defined in Definition \ref{def:C}) holds. Applying Lemma \ref{lem:L2_UB_any_budget}, we obtain
\als{
L_{k,n} %&= \sum_x \hat\pi_k(x) \sum_y (P_k(x,y) - \widehat P_{k,n}(x,y))^2 \\
&\le
\frac{2\beta}{T_{k,n}}\sum_{x:T_{k,x,n}>0} G_k(x)
+ \frac{2\sqrt{2}}{3}\frac{\beta^{3/2}\sqrt{S}}{T_{k,n}}\sum_x \frac{T_{k,x,n}^{3/2}\sqrt{G_k(x)}}{(T_{k,x,n}+\alpha S)^2} +
\frac{S\beta^2}{9T_{k,n}}\sum_x \frac{T_{k,x,n}}{(T_{k,x,n} + \alpha S)^2} \\
&\le
\frac{2\beta}{T_{k,n}}\sum_{x:T_{k,x,n}>0} G_k(x)
+ \frac{2\sqrt{2}}{3}\frac{\beta^{3/2}}{T_{k,n}}\sqrt{S\sum_{x:T_{k,x,n}>0} G_k(x)}\sqrt{\sum_{x:T_{k,x,n}>0} \frac{1}{T_{k,x,n}+\alpha S}} \\
&+
\frac{S\beta^2}{9T_{k,n}}\sum_{x:T_{k,x,n}>0} \frac{1}{T_{k,x,n} + \alpha S} \\
&=
\left(\sqrt{\frac{2\beta}{T_{k,n}}\sum_{x:T_{k,x,n}>0} G_k(x)} + \sqrt{\frac{S\beta^2}{9T_{k,n}}\sum_{x:T_{k,x,n}>0} \frac{1}{T_{k,x,n} + \alpha S}}\right)^2
\, ,
}
where we have used Cauchy-Schwarz in the second inequality. Introducing
\als{
A_{1,k} := \frac{S\beta^2}{9T_{k,n}}\sum_{x:T_{k,x,n}>0} \frac{1}{T_{k,x,n} + \alpha S} \quad\hbox{and}\quad    A_{2,k} := \frac{2\beta}{T_{k,n}}\sum_{x:T_{k,x,n}>0} G_k(x)  \, ,
}
we provide upper bounds on $A_{1,k}$ and $A_{2,k}$ in the following lemmas:

\begin{lemma}
\label{lem:Tk_LB_any_budegt_sqrt}
On the event $C$, it holds for any chain $k$ and any $n$:
\als{
A_{1,k} \le \frac{0.175 KS^2\beta^2}{n-K} \, .
}
\end{lemma}

\begin{lemma}
\label{lem:Tk_LB_any_budget_linear}
Assume that the event $C$ holds. Then for any chain $k$ and $n$:
\als{
A_{2,k} \le \frac{288KS^2\beta^2}{n-2K} + \frac{550K^2S^2\beta^2}{(n-2K)^2} \, .
}
\end{lemma}

Applying Lemmas \ref{lem:Tk_LB_any_budegt_sqrt} and \ref{lem:Tk_LB_any_budget_linear} gives
\als{
L_{k,n}
&\le
(\sqrt{A_{1,k}} + \sqrt{A_{2,k}})^2 \le \frac{KS^2\beta^2}{n-2K}\left(\sqrt{288 + \frac{550K}{n-2K}} + \sqrt{0.175}\right)^2\\
&\le KS^2\beta^2\left(\frac{304}{n-2K} + \frac{564K}{(n-2K)^2}\right) \, ,
}
where we have used
\als{
\left(\sqrt{288 + \frac{550K}{n-2K}} + \sqrt{0.175}\right)^2 \le 289 + \frac{550K}{n-2K} + 2\sqrt{0.175}\sqrt{288 +  \frac{550K}{n-2K}} \le 304 + \frac{564K}{n-2K}\, .
}
Finally, using the inequality $(n-2K)^{-1}\le n^{-1} + 4Kn^{-2}$ valid for $n\ge 4K$, and noting that the event $C$ holds with a probability higher than $1-\delta$, we get the desired bound on the loss.
\ep

\subsection{Proof of Lemma \ref{lem:Tk_LB_any_budegt_sqrt}}
Assume that $C$ holds.
We claim that there exists a chain $j$ such that $T_{j,n}\ge \frac{n}{K}$. We show this claim by contradiction: If for all $j$, $T_{j,n}<\frac{n}{K}$, then $\sum_{j=1}^K T_{j,n} < n$, which is a contradiction.

Now, consider a chain $j$ such that $T_{j,n}\ge \frac{n}{K}$. Let $t+1\le n$ be the last time that it has been sampled. Hence, $T_{j,t+1}=T_{j,n}$ and $T_{j,t}=T_{j,n}-1\ge \frac{n}{K}-1$. Applying Lemma \ref{lem:index_Sq_UB_LB_any_budget} for chain $j$, it follows that on the event $C$,
\als{
b_{j,t+1} &\le \frac{2\beta}{T_{j,t}}\sum_{x:T_{j,x,t}>0} G_j(x)
+ \frac{13\beta^{3/2}\sqrt{S}}{T_{j,t}}\sum_{x:T_{j,x,t}>0} \sqrt{\frac{G_j(x)}{T_{j,x,t} + \alpha S}}
+ \frac{39\beta^2S}{T_{j,t}}\sum_{x:T_{j,x,t}>0} \frac{1}{T_{j,x,t} + \alpha S} \\
&\le
\frac{2\beta}{T_{j,t}}\sum_{x:T_{j,x,t}>0} G_j(x)
+ \frac{13\beta^{3/2}\sqrt{S}}{T_{j,t}} \sqrt{\sum_{x:T_{j,x,t}>0} G_j(x)}\sqrt{\sum_{x:T_{j,x,t}>0} \frac{1}{T_{j,x,t} + \alpha S}} \\
&+ \frac{39\beta^2 S}{T_{j,t}}\sum_{x:T_{j,x,t}>0} \frac{1}{T_{j,x,t} + \alpha S} \\
&\le
\frac{K}{n-K}\left(2\beta\sum_x G_j(x)
+ 13\beta^{3/2} \sqrt{S\sum_x G_j(x)}\sqrt{\sum_x \frac{1}{1 + \alpha S}}
+ 39\beta^2S\sum_x \frac{1}{1 + \alpha S} \right)\\
&\le
\frac{K}{n-K}\left(2\beta \sum_x G_j(x) + 12S\beta^{3/2}\sqrt{\sum_x G_j(x)} + 30\beta^2S^2\right) \, .
}
Noting that $\sum_x G_j(x) \le S-1$, we get
\als{
b_{j,t+1}
&\le
\frac{K}{n-K}\left(2\beta (S-1) + 12S\beta^{3/2}\sqrt{S-1} + 30\beta^2S^2\right) \le \frac{44KS^2 \beta^2}{n-K}\,,
}
where we have used that $S\ge 2$. Note that for any chain $i$, by the definition of $b_{i,t+1}$,
$$
b_{i,t+1} \ge \frac{28\beta^2 S}{T_{i,t}} \sum_{x}\frac{\indic \{T_{i,x,t}>0\}}{T_{i,x,t}+\alpha S} \ge  \frac{28\beta^2 S}{T_{i,n}} \sum_{x}\frac{\indic \{T_{i,x,n}>0\}}{T_{i,x,n}+\alpha S} \, .
$$
Furthermore, since $j$ is played at time $t$, it holds that for any chain $i\neq j$, $b_{i,t+1}\le b_{j,t+1}$, so that for any chain $i$,
\als{
b_{j,t+1}\ge b_{i,t+1} \ge \frac{28\beta^2 S}{T_{i,n}}\sum_{x}\frac{\indic \{T_{i,x,n}>0\}}{T_{i,x,n}+\alpha S} \, .
}
Thus, combining this with the upper bound on $b_{j,t+1}$ leads to the desired results.
\ep

\subsection{Proof of Lemma \ref{lem:Tk_LB_any_budget_linear}}
The proof borrows some ideas from the proof of Lemma 1 in \cite{carpentier2011upper}.
Consider a chain $j$ that is sampled at least once after initialization, and let $t+1(>2K)$ be the last time it was sampled. Hence, $T_{j,t}=T_{j,n}-1$ and $T_{j,t+1}=T_{j,n}$. Moreover, let $X_{t+1}$ be the observed state of $j$ at $t+1$. Then, $T_{j,X_{t+1},t} = T_{j,X_{t+1},n} - 1$ and $T_{j,X_{t+1}, t+1}=T_{j,X_{t+1}, n}$, whereas for all $x\ne X_{t+1}$, $T_{j,x,t} = T_{j,x,t+1} = T_{j,x,n}$. We thus have,
$T_{j,x,t} \ge T_{j,x,n}-1$ for all $x\in \cS$.

By the design of the algorithm, for any chain $k$, $b_{k,t+1}\le b_{j,t+1}$. Using Lemma \ref{lem:index_Sq_UB_LB_any_budget} yields
\als{
\frac{2\beta}{T_{k,n}}\sum_x G_k(x) &\le \frac{2\beta}{T_{j,n}-1}\sum_x G_j(x)  + \frac{13\beta^{3/2}\sqrt{S}}{T_{j,n}-1}\sum_x \sqrt{\frac{G_j(x)}{T_{j,x,t} + \alpha S}}
+ \frac{39\beta^2 S}{T_{j,n}-1}\sum_x \frac{1}{T_{j,x,t} + \alpha S} \\
&\le
\frac{2\beta}{T_{j,n}-1}\sum_x G_j(x)  + \frac{26\beta^{3/2}\sqrt{S}}{T_{j,n}-1}\sum_x \sqrt{\frac{G_j(x)}{T_{j,x,n} + \alpha S}} \\
&+ \frac{156\beta^2 S}{T_{j,n}-1}\sum_x \frac{1}{T_{j,x,n} + \alpha S}\, ,
}
where in the second line we have used that for $\alpha = \frac{1}{3S}$ and $T_{j,x,n}\ge 1$,
\als{
T_{j,x,t} + \alpha S \ge T_{j,x,n} - 1 + \alpha S \ge \frac{T_{j,x,n} + \alpha S}{4}\, .
}
The above holds for any chain $k$, and any chain $j$ that is sampled at least once after the initialization (hence, $T_{j,n}>2$). Summing over such choices of $j$ gives
\als{
\frac{2\beta}{T_{k,n}}\sum_x G_k(x)&\sum_{j:T_{j,n}>2} (T_{j,n}-1)  \\
&\le 2\beta \sum_j \sum_x G_j(x) + 26\beta^{3/2}\sqrt{S}\sum_j \sum_{x} \sqrt{\frac{G_j(x)}{T_{j,x,n}+\alpha S}} + 156\beta^2 S\sum_j \sum_x \frac{1}{T_{j,x,n} + \alpha S} \\
&\le
2\beta \Lambda + 26\beta^{3/2}\sqrt{S\Lambda} \sqrt{\sum_j \sum_x \frac{1}{T_{j,x,n}+\alpha S}} + 156\beta^2 S\sum_j \sum_x \frac{1}{T_{j,x,n} + \alpha S}
\, ,
}
where we have used Cauchy-Schwarz in the last inequality, and that $\sum_j \sum_x G_j(x) = \Lambda$. Noting that $\sum_{j:T_{j,n}>2} (T_{j,n}-1) \ge n-2K$ yields
\als{
\frac{2\beta}{T_{k,n}}\sum_x G_k(x)
&\le
\frac{2\beta \Lambda}{n-2K} + \frac{26\beta^{3/2}\sqrt{\Lambda}}{n-2K} \sqrt{\sum_j \sum_x \frac{S}{T_{j,x,n}+\alpha S}} + \frac{156\beta^2}{n-2K}\sum_j \sum_x \frac{S}{T_{j,x,n} + \alpha S}
%&\le
%\frac{2}{n-2K}\left(\sqrt{\beta\Lambda} + \max\Big(\frac{c_3}{\sqrt{2\beta c_4}} ,\sqrt{c_4}\Big)\sqrt{\sum_j\sum_x \frac{S}{T_{j,x,n} + \alpha S}} \right)^2 \\
%&\le
%\frac{4}{n-2K}\left(\beta\Lambda + \max\Big(\frac{c^2_3}{2\beta c_4} ,c_4\Big)\sum_j\sum_x \frac{S}{T_{j,x,n} + \alpha S}\right)
\, .
}
By Lemma \ref{lem:Tk_LB_any_budegt_sqrt}, $\sum_x \frac{S}{T_{j,x,n}+\alpha S} \le \frac{1.58KS^2}{n-K}T_{j,n}$ for any chain $j$, which gives
\als{
\frac{2\beta}{T_{k,n}}\sum_x G_k(x)
&\le
\frac{2\beta \Lambda}{n-2K} + \frac{33\beta^{3/2}\sqrt{\Lambda}}{n-2K} \sqrt{\frac{KS^2}{n-K}\sum_j T_{j,n}} + \frac{263KS^2\beta^2}{(n-2K)(n-K)}\sum_j T_{j,n} \\
&\le
\frac{2\beta KS}{n-2K} + \frac{33\beta^{3/2}KS^{3/2}}{n-2K}\sqrt{\frac{n}{n-2K}} + \frac{263KS^2\beta^2n}{(n-2K)^2} \\
&\le
\frac{288KS^2\beta^2}{n-2K} + \frac{550K^2S^2\beta^2}{(n-2K)^2}
\, ,
}
where we have used $\sum_j T_{j,n} = n$, $\Lambda\le KS$, $S\ge 2$, and $\sqrt{\frac{n}{n-2K}}\le 1 + \frac{K}{n-2K}$.
\ep

\section{Proof of Theorem \ref{thm:loss_squared_suff_budget}}\label{sec:proof_suff_sample_regime}

Let $\delta\in (0,1)$ and $n\ge \text{\ncut}$. To control the loss in this case, we first state the following result for the concentration of empirical state distribution $\hat\pi_{k,n}$.

\begin{lemma}[{Concentration of Empirical State Distributions}]
\label{lem:inv_stat_dist_concent}
Assume that event $C$ holds and $n\ge \text{\ncut}$. Then, for any chain $k$ and state $x$, $\hat \pi_{k,n}(x)^{-1} \le 2\pi_k(x)^{-1}$ with probability at least $1-\delta$.
\end{lemma}

Recalling that $\hat\pi_{k,n}(x) = \frac{T_{k,x,n}}{T_{k,n}}$ for all $x\in \cS$, on the event $C$ (defined in Definition \ref{def:C}), we have by Lemma \ref{lem:L2_UB_any_budget} and Lemma \ref{lem:inv_stat_dist_concent},
\als{
L_{k,n}&
\le
 \frac{2\beta}{T_{k,n}}\sum_{x} G_k(x) +  \frac{2\sqrt{2}}{3}\frac{\beta^{3/2}\sqrt{S}}{T_{k,n}^{3/2}}\sum_{x} \sqrt{\frac{G_k(x)}{\hat\pi_{k,n}(x)}} + \frac{S\beta^2}{9T_{k,n}^2}\sum_{x} \frac{1}{\hat \pi_{k,n}(x)} \\
&\le
\frac{2\beta}{T_{k,n}}\sum_{x} G_k(x) +  \frac{4\beta^{3/2}\sqrt{S}}{3T_{k,n}^{3/2}}\sum_{x} \sqrt{\frac{G_k(x)}{\pi_{k}(x)}}
+  \frac{2SH_k \beta^2}{9T_{k,n}^2} \\
&\le
\frac{2\beta}{T_{k,n}}\sum_{x} G_k(x)+  \frac{4\beta^{3/2}}{3T_{k,n}^{3/2}}\sqrt{SH_k\sum_{x}G_k(x)}
+  \frac{2SH_k \beta^2}{9T_{k,n}^2}\, ,
}
with probability at least $1-\delta$, where the last step follows from the Cauchy-Schwarz inequality.

To control the right-hand side of the above, we first provide an upper bound on $\frac{2\beta}{T_{k,n}}\sum_x G_k(x)$ assuming that the event $C$ holds:

\begin{lemma}
\label{lem:G_over_T_UB}
Assume that the event $C$ holds. Then, for any chain $k$ and $n\ge \text{\ncut}$, it holds that
\begin{align*}
\frac{2\beta}{T_{k,n}}\sum_x G_k(x) &\le \frac{A_1}{n} + \frac{A_2}{n^{3/2}}  + \frac{A_3}{n^2}  + \widetilde\Ocal(n^{-5/2}) \, ,
\end{align*}
with probability at least $1-\delta$, where
\als{
A_1 = 2\beta \Lambda, \quad A_2 = 150\beta^{3/2} K\sqrt{S\Lambda H_{\max}}, \quad A_3 = \frac{3912KSH_{\max}\beta^2}{\eta_{\min}}\, .
}
\end{lemma}

Applying Lemma \ref{lem:G_over_T_UB}, and noting $\PP(C) \ge 1-\delta$ (see Lemma \ref{lem:P_event_C}), we obtain the following bound on $L_{k,n}$, which holds with probability greater than $1-2\delta$:
%\als{
%\frac{\beta}{T_{k,n}}\sum_{x} G_k(x) &\le \frac{A_1}{n} + \frac{A_2}{n^{3/2}} + \frac{A_3}{n^2}\, .
%}
%Then,
\als{
L_{k,n}
&\le
\frac{2\beta}{T_{k,n}}\sum_{x} G_k(x)+  \frac{4\beta^{3/2}}{3T_{k,n}^{3/2}}\sqrt{SH_k\sum_{x}G_k(x)}
+  \frac{2SH_k \beta^2}{9T_{k,n}^2} \\
&\le
 \frac{2\beta}{T_{k,n}}\sum_{x} G_k(x)+  \left(\frac{2\beta}{T_{k,n}}\sum_{x} G_k(x)\right)^{3/2} \frac{0.48\sqrt{SH_k}}{\sum_{x}G_k(x)}
+  \frac{2SH_k}{9(\sum_{x}G_k(x))^2} \left(\frac{\beta}{T_{k,n}}\sum_{x} G_k(x)\right)^2 \\
&\le
  \frac{A_1}{n} + \frac{A_2}{n^{3/2}} + \frac{A_3}{n^2} +  \left(\frac{A_1}{n} + \frac{A_2}{n^{3/2}} + \frac{A_3}{n^2}\right)^{3/2} \frac{0.48\sqrt{SH_k}}{\sum_{x}G_k(x)} \\
&+  \frac{2SH_k}{9(\sum_{x}G_k(x))^2} \left(\frac{A_1}{n} + \frac{A_2}{n^{3/2}} + \frac{A_3}{n^2}\right)^2  + \widetilde\cO(n^{-5/2})\\
&\stackrel{(\text{a})}\le
  \frac{A_1}{n} + \frac{A_2}{n^{3/2}} + \frac{A_3}{n^2} +  \frac{0.84\sqrt{SH_k}}{\sum_{x}G_k(x)}\left(\frac{A_1^{3/2}}{n^{3/2}} 
  + \frac{A_2^{3/2}}{n^{9/4}} + \frac{A_3^{3/2}}{n^3}\right)\\
&+  \frac{4SH_k}{9(\sum_{x}G_k(x))^2} \left(\frac{A_1^2}{n^2} + \frac{A_2^2}{n^{3}} + \frac{A_3^2}{n^4}\right) + \widetilde\cO(n^{-5/2})\\
&\le
  \frac{A_1}{n} + \frac{1}{n^{3/2}}\left(A_2 + \frac{0.84\sqrt{SH_k}}{\sum_{x}G_k(x)}A_1^{3/2}\right)
  + \widetilde\cO(n^{-2}) \\
  %&+
 % \frac{1}{n^{2}}\left(A_3 + \frac{0.84\sqrt{SH_k}}{\sum_{x}G_k(x)}\left(A_2^{3/2} + \frac{A_3^{3/2}}{4K}\right) + \frac{4SH_k}{9(\sum_{x}G_k(x))^2} \left(A_1^2 + \frac{A_2^2}{4K} + \frac{A_3^2}{16K^2}\right) \right)\\
  &\le
  \frac{2\beta \Lambda}{n} + \frac{\beta^{3/2}}{n^{3/2}}\left( 150K\sqrt{S\Lambda H_{\max}} + \frac{2.4\sqrt{SH_k\Lambda}}{\eta_k}\right)  + \widetilde\cO(n^{-2})
  %\frac{1}{n^{2}}\left(A_3 + \frac{\sqrt{3}\square\sqrt{SH_k}}{\sum_{x}G_k(x)}\left(A_2^{3/2} + \frac{A_3^{3/2}}{4K}\right) + \frac{2SH_k}{(\sum_{x}G_k(x))^2} \left(A_1^2 + \frac{A_2^2}{4K} + \frac{A_3^2}{16K^2}\right) \right)\\
\, ,
}
where (a) follows from the fact that  for positive numbers  $a_i, i=1,\ldots, m$, we have by Jensen's inequality,
\als{
\Big(\frac{1}{m}\sum_{i=1}^m a_i\Big)^{3/2} \le \frac{1}{m}\sum_{i=1}^m a_i^{3/2}\, .
}
so that $(\sum_{i=1}^m a_i)^{3/2} \le \sqrt{m}\sum_{i=1}^m a_i^{3/2}$. Finally, taking the maximum over $k$ completes the proof.
\ep

\subsection{Proof of Lemma \ref{lem:inv_stat_dist_concent}}
By Lemma \ref{lem:stat_concent_Hsu}, we have for all chains $k$ and all $x\in \Scal$,
\als{
|\hat \pi_{k,n}(x) - \pi_k(x)| &\le \xi_{k,x,n}:= \sqrt{\frac{8\pi_k(x)\epsilon_k}{T_{k,n}}} + \frac{20\epsilon_k}{T_{k,n}} \, ,
}
with probability at least $1-\delta$, where $\epsilon_k:=\frac{1}{\gamma_k}\log\Big(\frac{K}{\delta}\sqrt{\frac{2}{\underline{\pi}_k}}\Big)$. It is easy to verify that if $T_{k,n}\ge \frac{96\epsilon_k}{\underline{\pi}_k}$, then
$\xi_{k,x,n}\le \pi_k(x)/2$, so that for all $k$ and all $x$,
\als{
\frac{1}{\hat\pi_{k,n}(x)} &= \frac{1}{\pi_k(x)} + \frac{\pi_k(x) - \hat \pi_{k,n}(x)}{\hat \pi_{k,n}(x)\pi_k(x)} \\
 &\le \frac{1}{\pi_k(x)} + \frac{\xi_{k,x,n}}{\pi_k(x)(\pi_k(x) -\xi_{k,x,n}) } \\
 &\le \frac{2}{\pi_k(x)} \, ,
}
with probability at least $1-\delta$.

It remains to show that if $n\ge \text{\ncut}$, we have $T_{k,n}\ge \frac{96\epsilon_k}{\underline{\pi}_k}$. Indeed, when $C$ occurs, as a consequence of Lemma \ref{lem:Tk_LB_any_budegt_sqrt}, one has
\als{
\frac{S\beta^2}{9T_{k,n}}\sum_{x:T_{k,x,n}>0} \frac{1}{T_{k,x,n}+\alpha S} \le \frac{0.175KS^2\beta^2}{n-K}\, ,
}
with probability at least $1-\delta$. Using the trivial bound $T_{k,x,n}\le T_{k,n}$, it follows that
\als{
\frac{S^2\beta^2}{9T_{k,n}(T_{k,n}+\alpha S)} \le \frac{0.175KS^2\beta^2}{n-K}\, ,
}
so that
$$
T_{k,n} \ge \sqrt{\frac{n-K}{1.575K}} - \frac{1}{3} \ge 0.56\sqrt{\frac{n}{K}}-\frac{1}{3} \ge 0.327\sqrt{\frac{n}{K}} \, .
$$
%$$
%T_{k,n} \ge 0.77\sqrt{\frac{n-K(1+\alpha S)}{K}} \ge 0.77\sqrt{\frac{n}{2K}} \ge 0.5\sqrt{\frac{n}{K}} \, .
%$$
with probability greater than $1-\delta$. Putting together, we deduce that if $n$ satisfies $0.327\sqrt{\frac{n}{K}}\ge \frac{96\epsilon_k}{\underline{\pi}_k}$, we have $\xi_{k,x,n}\le \pi_k(x)/2$, and the lemma follows.

Moreover, when the chain $k$ is non-reversible, we may use \cite[Theorem~3.4]{paulin2015concentration} (instead of Lemma \ref{lem:Tk_LB_any_budegt_sqrt}), and follow the exact same lines as above to deduce that if
$n\ge K\max_k \Big(\frac{300}{\gamma_{\textsf{ps},k}\underline{\pi}_k}\log\Big(\frac{2K}{\delta}\sqrt{\underline{\pi}_k^{-1}}\Big)\Big)^2$, the assertion of the lemma follows.
\ep

%In the following lemma, we present a lower bound on $T_{k,n}$ for any chain $k$:
%
%\begin{lemma}
%\label{lem:LB_Tkn_ii}
%Assume that event $C$ holds. Then, for any chain $k$,
%\als{
%T_{k,n} \ge 0.29\sqrt{n/K},
%}
%\end{lemma}
%
%\bp
%Consider a chain $k$ and time $t$. Recall from the proof of Lemma \ref{lem:G_over_T_UB} that there exists a chain $j$ that is sampled for the last time at time $t+1$. So using the upper bound in Lemma \ref{lem:index_UB_LB} and that $T_{j,n}\ge 2$, we get
%\als{
%b_{k,t} &\le b_{j,t} \\
%&\le \frac{\beta}{T_{j,n}-1}\sum_{x} G_{j}(x) +  \frac{8\beta\sqrt{S}}{T_{j,n}-1} \sum_x\sqrt{\frac{\sum_x G_j(x)}{T_{j,x,t}}} + \frac{17\beta^2S}{T_{j,n}-1}\sum_x \frac{1}{T_{j,x,t}} \\
%&\le \frac{2\beta S}{T_{j,n}} +  \frac{8\sqrt{2}\beta S^{3/2}}{T_{j,n}} + \frac{17\beta^2S^2}{T_{j,n}} \\
%&\le \frac{26\beta^2S^2}{T_{j,n}}   \, .
%}
%where we used $S\ge 2$. On the other hand,
%\als{
%b_{k,t} &\ge \frac{9\beta^2 S}{4T_{k,t}}\sum_x \frac{1}{T_{k,x,t}-1} \ge \frac{9\beta^2 S}{4T_{k,t}}\sum_x \frac{1}{T_{k,x,t}} \, .
%}
%By Jensen's inequality, $\sum_{x} T_{k,x,t}^{-1} \ge S(\sum_x T_{k,x,t})^{-1} = ST_{k,t}^{-1}$, so that
%\als{
%b_{k,t} &\ge \frac{9\beta^2 S^2}{4T_{k,t}^2} \ge \frac{9\beta^2 S^2}{4T_{k,n}^2}\, .
%}
%Putting these bounds on $b_{k,t}$ together gives
%\als{
%\frac{9\beta^2 S^2}{4T_{k,n}^2}\sum_{j} T_{j,n} &\le 26\beta^2S^2K.
%}
%Noting that $\sum_j T_{j,n}=n$ completes the proof.
%\ep

\subsection{Proof of Lemma \ref{lem:G_over_T_UB}}
The proof borrows some ideas from the proof of Lemma 1 in \cite{carpentier2011upper}.
Consider a chain $j$ that is sampled at least once after initialization, and let $t+1(>2K)$ be the last time it was sampled. Hence, $T_{j,t}=T_{j,n}-1$ and $T_{j,t+1}=T_{j,n}$. Moreover, let $X_{t+1}$ be the observed state of $j$ at $t+1$. Then, $T_{j,X_{t+1},t} = T_{j,X_{t+1},n} - 1$ and $T_{j,X_{t+1}, t+1}=T_{j,X_{t+1}, n}$, whereas for all $x\ne X_{t+1}$, $T_{j,x,t} = T_{j,x,t+1} = T_{j,x,n}$. We thus have,
$T_{j,x,t} \ge T_{j,x,n}-1$ for all $x\in \cS$.

By the design of the algorithm, for any chain $k$, $b_{k,t+1}\le b_{j,t+1}$. Applying Lemma \ref{lem:index_Sq_UB_LB_any_budget} gives
\begin{align*}
   \frac{2\beta}{T_{k,t}}\sum_{x} G_k(x) &\le
    \frac{2\beta}{T_{j,t}}\sum_{x} G_j(x) + \frac{c_3\sqrt{S}}{T_{j,t}}\sum_x \sqrt{\frac{G_j(x)}{T_{j,x,t} + \alpha S}} + \frac{c_4S}{T_{j,t}}\sum_x \frac{1}{T_{j,x,t} + \alpha S} \\
   &\le
    \frac{2\beta}{T_{j,n}-1}\sum_{x} G_j(x) + \frac{2c_3\sqrt{S}}{T_{j,n}-1}\sum_x \sqrt{\frac{G_j(x)}{T_{j,x,n} + \alpha S}} + \frac{4c_4S}{T_{j,n}-1}\sum_x \frac{1}{T_{j,x,n} + \alpha S}
   \, ,
\end{align*}
where $c_3 = 13\beta^{3/2}$ and $c_4 = 39\beta^2$, and where
where in the second line we have used that for $\alpha = \frac{1}{3S}$ and $T_{j,x,n}\ge 1$
\als{
T_{j,x,t} + \alpha S \ge T_{j,x,n} - 1 + \alpha S \ge \frac{T_{j,x,n} + \alpha S}{4}\, .
}
Now, applying Lemma \ref{lem:inv_stat_dist_concent}  and using $T_{k,t}\le T_{k,n}$ yield
\als{
\frac{2\beta}{T_{k,n}}\sum_{x} G_{k}(x) &\le \frac{2\beta}{T_{j,n}-1}\sum_{x} G_{j}(x) +  \frac{c_3\sqrt{8S}}{(T_{j,n}-1)T_{j,n}^{1/2}}\sum_x \sqrt{\frac{G_j(x)}{\pi_j(x)}} + \frac{8c_4S}{(T_{j,n}-1)T_{j,n}}\sum_x \frac{1}{\pi_j(x)}\\
%&\le
% \frac{2\beta}{T_{j,n}-1}\sum_{x} G_{j}(x) + \frac{2c_3\sqrt{S}L_j}{(T_{j,n}-1)T_{j,n}^{1/2}} + \frac{4c_4SH_j}{(T_{j,n}-1)T_{j,n}} \\
&\le
 \frac{1}{T_{j,n}-1}\left(2\beta\sum_{x} G_{j}(x) + \frac{\sqrt{8}c_3}{T_{j,n}^{1/2}}\sqrt{SH_j\sum_x G_j(x)} + \frac{8c_4SH_j}{T_{j,n}}\right)
   \, .
}
Note that the above relation is valid for any $k$, and any $j$ that is sampled after initialization (i.e., $T_{j,n}>2$).
Summing over such choices of $j$ gives
\als{
   \sum_{j:T_{j,n}>2}&\frac{2\beta \sum_x G_k(x)}{T_{k,n}}(T_{j,n}-1) \le \sum_{j:T_{j,n}>2}\left(2\beta \sum_x G_{j}(x) +  \frac{c_3\sqrt{8SH_j\sum_x G_j(x)}}{T_{j,n}^{1/2}} + \frac{8c_4 SH_j}{T_{j,n}} \right)\, .
}
Noting that $\sum_{j:T_{j,n}>2}(T_{j,n}-1)\ge n-2K$, we have
\begin{align}
   \frac{2\beta}{T_{k,n}}\sum_x G_k(x)
   &\le
    \frac{1}{n-2K}\sum_{j}\left(2\beta \sum_x G_{j}(x) +  \frac{c_3\sqrt{8SH_j\sum_x G_j(x)}}{T_{j,n}^{1/2}} + \frac{8c_4SH_j}{T_{j,n}}\right) \nonumber\\
   &\le
    \frac{2\beta \Lambda}{n-2K} + \frac{c_3\sqrt{8S}}{n-2K}\sum_j \sqrt{H_j \frac{\sum_x G_j(x)}{T_{j,n}}} + \frac{8c_4S}{n-2K}\sum_j \frac{H_j}{T_{j,n}}
    \, .
   \label{eq:ratio_p_k_1}
\end{align}

To carefully control the right-hand side of the last inequality, we use the following lemma:

\begin{lemma}
\label{lem:sqrt_Gk_over_Tk_UB}
Under the same assumption of Lemma \ref{lem:G_over_T_UB}, we have for any chain $j$,
\begin{align*}
\frac{\beta}{T_{j,n}} &\le \frac{\beta}{\eta_{\min}(n-2K)} + \frac{c_3}{\eta_{\min}^2(n-2K)^{3/2}}\sqrt{2SH_{\max}/\Lambda} + \frac{4c_4SH_{\max}}{\Lambda\eta_{\min}^3(n-2K)^2}\, , \nonumber  \\
\sqrt{\frac{2\beta}{T_{j,n}} \sum_x G_j(x)} &\le \sqrt{\frac{2\beta\Lambda}{n-2K}} + \frac{18\beta\sqrt{SH_{\max}}}{\eta_{\min}(n-2K)}    \, .
\end{align*}
\end{lemma}

Now, applying Lemma \ref{lem:sqrt_Gk_over_Tk_UB} yields
\begin{align*}
\frac{2\beta}{T_{k,n}}\sum_{x} G_k(x) &\le \frac{2\beta \Lambda}{n-2K} + \frac{37\beta\sqrt{SH_{\max}}}{n-2K}\sum_j
\left( \sqrt{\frac{2\beta\Lambda}{n-2K}} + \frac{18\beta\sqrt{SH_{\max}}}{\eta_{\min}(n-2K)}  \right) \\
&+ \frac{312\beta S H_{\max}}{n-2K} \sum_j \left( \frac{\beta}{\eta_{\min}(n-2K)} + \frac{c_3}{\eta_{\min}^2(n-2K)^{3/2}}\sqrt{2SH_j/\Lambda} + \frac{4c_4SH_j}{\Lambda\eta_{\min}^3(n-2K)^2}  \right) \\
&\le \frac{2\beta \Lambda}{n-2K} + \frac{53\beta^{3/2} K\sqrt{SH_{\max}\Lambda}}{(n-2K)^{3/2}} + \frac{978KSH_{\max}\beta^2}{\eta_{\min}(n-2K)^2} + \widetilde\Ocal((n-2K)^{-5/2})
\, .
\end{align*}
%where we used $n-2K \ge 2K$, and $S\ge 2$, $K\ge 2$.

Finally, using the inequality $(n-2K)^{-1}\le n^{-1} + 4K n^{-2}$ and $n-2K\ge n/2$ valid for all $n\ge 4K$, we get the desired result:
\begin{align*}
\frac{2\beta}{T_{k,n}} \sum_x G_k(x) &\le \frac{2\beta \Lambda}{n} + \frac{150\beta^{3/2} K\sqrt{S\Lambda H_{\max}}}{n^{3/2}}  + \frac{3912KSH_{\max}\beta^2}{\eta_{\min}n^2}  + \widetilde\Ocal(n^{-5/2})\, .
\end{align*}
\ep

\subsection{Proof of Lemma \ref{lem:sqrt_Gk_over_Tk_UB}}
The proof borrows some ideas from the proof of Lemma 1 in \cite{carpentier2011upper}.
Consider a chain $j$ that is sampled at least once after initialization, and let $t+1(>2K)$ be the last time it was sampled. Hence, $T_{j,t}=T_{j,n}-1$. Using the same arguments as in the beginning of the proof of Lemma \ref{lem:G_over_T_UB}, we have on the event $C$,
\begin{align}
\label{eq:Tk_Tj_ii}
\frac{2\beta}{T_{k,n}}\sum_{x} G_{k}(x) &\le \frac{1}{T_{j,n}-1}\left(2\beta\sum_{x} G_{j}(x) +   c_3\sqrt{\frac{8SH_j\sum_x G_j(x)}{T_{j,n}}} + \frac{8c_4SH_j}{T_{j,n}}\right)
   \, .
\end{align}
Note that (\ref{eq:Tk_Tj_ii}) is valid for any $k$, and any $j$ that is sampled after initialization.

Now consider a chain $j$ such that $T_{j,n}-2\ge \eta_j(n-2K)$. In other words, $j$ is over-sampled (w.r.t.~budget $n-2K$). In particular, $j$ is sampled at least once after initialization. Hence, using (\ref{eq:Tk_Tj_ii}) and noting that $T_{j,n}\ge \eta_j(n-2K) + 2$, we obtain
\begin{align}
   \frac{2\beta}{T_{k,n}}\sum_x G_k(x)
   &\le
    \frac{1}{\eta_j(n-2K)}\left(
   2\beta \sum_x G_j(x) +
    c_3\sqrt{\frac{8SH_j\sum_x G_j(x)}{\eta_j (n-2K)}} + \frac{8c_4SH_j}{\eta_j(n-2K)}    \right) \nonumber\\
    &\stackrel{\text{(a)}}\le
    \frac{2\beta \Lambda}{n-2K} +  \frac{c_3\sqrt{8S\Lambda H_j}}{\eta_j(n-2K)^{3/2}} + \frac{8c_4SH_j}{\eta_j^2(n-2K)^2} \, ,
   \label{eq:lem_1}
\end{align}
where (a) follows from the definition of $\eta_j$.
Multiplying both sides on $\frac{\eta_k}{2\Lambda}$ gives:
\begin{align}
\label{eq:inv_Tkn_Sq}
\frac{\beta}{T_{k,n}}
&\le
 \frac{\beta}{\eta_k(n-2K)} +  \frac{c_3\sqrt{2S\Lambda H_j}}{\Lambda \eta_k\eta_j(n-2K)^{3/2}} + \frac{4c_4SH_j}{\Lambda\eta_k\eta_j^2(n-2K)^2} \\
&\le
\frac{\beta}{\eta_{\min}(n-2K)} + \frac{c_3}{ \eta_{\min}^2(n-2K)^{3/2}}\sqrt{2SH_j/\Lambda} + \frac{4c_4SH_j}{\Lambda\eta_{\min}^3(n-2K)^2}\, , \nonumber
\end{align}
thus verifying the first statement of the lemma. To derive the second statement, we take square-root from both sides of (\ref{eq:lem_1}):
\begin{align*}
\sqrt{\frac{2\beta}{T_{k,n}}\sum_x G_k(x)}
&\le
\sqrt{
\frac{2\beta \Lambda}{n-2K} +  \frac{c_3 \sqrt{8S\Lambda H_j}}{\eta_j(n-2K)^{3/2}} + \frac{8c_4SH_j}{\eta_j^2(n-2K)^2}
}
\\
   &\le
   \sqrt{
\frac{2\beta \Lambda}{n-2K} +  \frac{c_3\sqrt{8S\Lambda H_j}}{\eta_j(n-2K)^{3/2}}} + \frac{\sqrt{8c_4SH_j}}{\eta_j(n-2K)}
 \\
    &\le
     \sqrt{\frac{2\beta\Lambda}{n-2K}} + \frac{\sqrt{c_3/\beta}\sqrt{SH_{\max}}}{\eta_{\min}(n-2K)} + \frac{\sqrt{8c_4SH_{\max}}}{\eta_{\min}(n-2K)} \, ,
\end{align*}
where the second and third inequalities respectively follow from  $\sqrt{a+b}\le\sqrt{a} + \sqrt{b}$ and  $\sqrt{a+b}\le \sqrt{a}+\frac{b}{2\sqrt{a}}$ valid for all $a,b>0$. Plugging $c_3 = 13\beta^{3/2}$ and $c_4 =39\beta^2$ into the last inequality verifies the second statement and concludes the proof.
\ep

\section{Asymptotic Analyses --- Proofs  of Lemma \ref{lem:oracle_loss} and Theorem \ref{thm:loss_squared_asymp}}\label{sec:proof_asymptotic}

\subsection{Proof  of Lemma \ref{lem:oracle_loss}}
%Gamma distribution with shape parameters $\alpha,\beta$: Mean $\frac{\alpha}{\beta}$. $X\sim \Gamma(\alpha,\beta)$, then $\EE[X]=\frac{\alpha}{\beta}$.

Consider a chain $k$, and let us denote by $\widetilde P_{k,n}$ the corresponding empirical estimator of $P_k$ (corresponding to $\alpha = 0$). That is, for all $(x,y)\in \cS^2$, $\widetilde P_{k,n}(x,y) = \frac{1}{T_{k,x,n}}\sum_{t=2}^{n} \indic\{X_{t-1} = x, X_{t}=y\}$. Further, let $\widetilde L_{k,n}$ denote the corresponding loss for chain $k$.
To prove the lemma, we first show that $\lim_{T_{k,n}\to\infty} T_{k,n}L_{k,n}  =  \lim_{T_{k,n}\to\infty} T_{k,n}\widetilde L_{k,n}$. 

To this end, we derive upper and lower bounds on $L_{k,n}$ in terms of $\widetilde L_{k,n}$. We have for all $(x,y)\in\cS^2$:
\als{
|(\widehat P_{k,n} - \widetilde P_{k,n})(x,y)| &= \bigg|\frac{\sum_{t=2}^{n} \indic\{X_{t-1} = x, X_{t}=y\}+\alpha}{T_{k,x,n}+\alpha S} - \frac{\sum_{t=2}^{n} \indic\{X_{t-1} = x, X_{t}=y\}}{T_{k,x,n}}\bigg| \\
&= \frac{\alpha}{T_{k,x,n}(T_{k,x,n}+\alpha S)} \bigg|T_{k,x,n} - S\sum_{t=2}^{n} \indic\{X_{t-1} = x, X_{t}=y\}\bigg| \\
&\le \frac{\alpha ST_{k,x,n}}{T_{k,x,n}(T_{k,x,n}+\alpha S)} \le \frac{\alpha S}{T_{k,x,n}}\, .
}
We therefore get, on the one hand,
\als{
L_{k,n} &\ge \sum_x \hat\pi_{k,n}(x) \|\widehat P_{k,n}(x,\cdot) - \widetilde P_{k,n}(x,\cdot)\|_2^2 + \widetilde L_{k,n} \ge \widetilde L_{k,n} \, ,
}
and on the other hand,
\als{
L_{k,n} &\le \sum_x \hat\pi_{k,n}(x) \|\widehat P_{k,n}(x,\cdot) - \widetilde P_{k,n}(x,\cdot)\|_2^2 + \sum_x \hat\pi_{k,n}(x) \|P_k(x,\cdot) - \widetilde P_{k,n}(x,\cdot) \|^2_2 \\
&+ 2\underbrace{\sum_x \hat\pi_{k,n}(x) \sum_y \big|\widehat P_{k,n}(x,y) - \widetilde P_{k,n}(x,y)\big|\big|\widetilde P_{k,n}(x,y) -  P_k(x,y)\big|}_{A} \\
&\le
\sum_x \hat\pi_{k,n}(x) \|\widehat P_{k,n}(x,\cdot) - \widetilde P_{k,n}(x,\cdot)\|_2^2 + \widetilde L_{k,n} +  2A\\
&\le \sum_x \frac{T_{k,x,n}}{T_{k,n}}\sum_y \Big(\frac{\alpha S}{T_{k,x,n}}\Big)^2 + \widetilde L_{k,n}  + 2A = \frac{\alpha^2 S^3}{T_{k,n}}\sum_x \frac{1}{T_{k,x,n}} + \widetilde L_{k,n} + 2A \, .
}
Furthermore, $A$ is upper bounded as follows:
\als{
A &\le \sum_x \hat\pi_{k,n}(x)\frac{\alpha S}{T_{k,x,n}} \sum_y \big|\widetilde P_{k,n}(x,y) -  P_k(x,y)\big| \\
&\le \sqrt{ \sum_x \sum_y\hat\pi_{k,n}(x)\frac{\alpha^2 S^2}{T_{k,x,n}^2} } \sqrt{\sum_x \hat\pi_{k,n}(x)\sum_y (\widetilde P_{k,n}(x,y) -  P_k(x,y))^2} \\
&= \sqrt{ \sum_x \frac{\alpha^2 S^3}{T_{k,n}T_{k,x,n}} } \sqrt{\widetilde L_{k,n}} \, ,
}
where we have used Cauchy-Schwarz in the second line. In summary, we have shown that
\als{
\widetilde L_{k,n}\le L_{k,n} \le \widetilde L_{k,n} + \frac{\alpha^2 S^3}{T_{k,n}}\sum_x \frac{1}{T_{k,x,n}} + 2\sqrt{ \sum_x \frac{\alpha^2 S^3}{T_{k,n}T_{k,x,n}} } \sqrt{\widetilde L_{k,n}} \, ,
}
so that
\als{
T_{k,n}\widetilde L_{k,n}\le T_{k,n}L_{k,n} \le T_{k,n}\widetilde L_{k,n} + \alpha^2 S^3\sum_x \frac{1}{T_{k,x,n}} + 2\sqrt{ \sum_x \frac{\alpha^2 S^3}{T_{k,x,n}} } \sqrt{T_{k,n}\widetilde L_{k,n}}  \, .
}
Taking the limit when $T_{k,n}\to\infty$ and noting the fact that when $T_{k,n}\to \infty$, by ergodicity, $T_{k,x,n}\to\infty$ for all $x\in \Scal$, we obtain
$\lim_{T_{k,n}\to\infty} T_{k,n}L_{k,n}  =  \lim_{T_{k,n}\to\infty} T_{k,n}\widetilde L_{k,n}$.

It remains to compute $\lim_{T_{k,n}\to\infty} T_{k,n}\widetilde L_{k,n}$. We have
\als{
\widetilde L_{k,n} &= \frac{1}{T_{k,n}}\sum_x T_{x,n} \sum_y (\widetilde P_{k,n}(x,y) - P_{k}(x,y))^2 \\
 &= \frac{1}{T_{k,n}}\sum_x \sum_y \underbrace{[\sqrt{T_{k,x,n}}(\widetilde P_{k,n}(x,y) - P_{k}(x,y))]^2}_{Z(x,y)} \, .
  }
 %Using $T_{k,n} = \frac{\sum_x G_k(x)}{\Lambda} n$, we have $\widetilde L_{k,n}\to_{n\to\infty} \Lambda/n$.
When $T_{k,n}\to \infty$, by ergodicity, we have $T_{k,x,n}\to\infty$ for all $x\in \Scal$. Therefore, by the central limit theorem, $\sqrt{T_{k,x,n}}(\widetilde P_{k,n}(x,y) - P_{k}(x,y))$ converges (in distribution) to a Normal distribution with variance $P_k(I-P_k)(x,y)$, and $Z(x,y)$ converges to a Gamma distribution with mean $P_k(I-P_k)(x,y)$. Hence, the mean of $\widetilde L_{k,n}$ would be  $\frac{1}{T_{k,n}}\sum_x\sum_y P_k(I-P_k)(x,y) = \frac{1}{T_{k,n}}\sum_{x} G_k(x)$,  thus completing the proof.
\ep

\subsection{Proof of  Theorem \ref{thm:loss_squared_asymp}}

Let $n$ be a budget such that $\sqrt{n}\ge \text{\ncut}$, and let $n_0:=\sqrt{n}$. The proof proceeds in two steps. We first consider the case with budget $n_0$, and show that at the end of this sub-problem, the index of each chain is well estimated. Then, we consider allocation in the second sub-problem. 

\paragraph{Step 1: Bounds on the index at $t=n_0$.}Fix a chain $k$.  
To derive upper and lower bounds on the index $b_{k,t}$ at $t=n_0$, we derive a lower bound on $T_{k,n_0}$. 
Recall that by Lemma \ref{lem:Tk_LB_any_budegt_sqrt}, we have with the choice $n=n_0$,
\als{
\frac{S\beta^2}{9T_{k,n_0}}\sum_{x:T_{k,x,n_0}>0} \frac{1}{T_{k,x,n_0}+\alpha S} \le \frac{0.187KS^2\beta^2}{n_0-K}\, ,
}
with probability at least $1-\delta$. Using the trivial bound $T_{k,x,n_0}\le T_{k,n_0}$, it follows that
\als{
\frac{S^2\beta^2}{9T_{k,n_0}(T_{k,n_0}+\alpha S)} \le \frac{0.175KS^2\beta^2}{n_0-K}\, ,
}
so that
$$
T_{k,n_0} \ge \sqrt{\frac{n_0-K}{1.575K}} - \frac{1}{3} \ge  \frac{1}{4}\sqrt{\frac{n_0}{K}} \ge \frac{n^{1/4}}{4\sqrt{K}}\, ,
$$
with probability greater than $1-\delta$.

Noting that $n_0\ge \text{\ncut}$, we apply Lemma \ref{lem:index_Sq_UB_LB_any_budget} and \ref{lem:inv_stat_dist_concent} to obtain
\als{
\frac{2\beta}{T_{k,n_0}}\sum_x G_k(x) \le b_{k,n_0+1} \le \frac{2\beta}{T_{k,n_0}}\sum_x G_k(x)
+ \frac{19\beta^{3/2}\sqrt{S}}{T_{k,n_0}^{3/2}}\sum_x \sqrt{\frac{G(x)}{\pi_k(x)}}
+ \frac{90\beta^2 S}{T^2_{k,n_0}}\sum_{x} \frac{1}{\pi_k(x)}\, ,
}
with probability at least $1-2\delta$.
Using Cauchy-Schwarz and recalling $H_k:=\sum_k \pi_k(x)^{-1}$, we obtain give
\als{
\frac{2\beta}{T_{k,n_0}}\sum_x G_k(x) \le b_{k,n_0+1} \le \frac{2\beta}{T_{k,n_0}}\sum_x G_k(x)
+ \frac{19\beta^{3/2}}{T_{k,n_0}^{3/2}}\sqrt{SH_k \sum_x G_k(x)}+ \frac{78\beta^2 SH_k}{T^2_{k,n_0}}\, ,
}
so that
\als{
\sum_x G_k(x) \le \frac{T_{k,n_0}}{2\beta}b_{k,n_0+1} \le \sum_x G_k(x)
+ \frac{10\beta^{1/2}}{T_{k,n_0}^{1/2}}\sqrt{SH_k \sum_x G_k(x)}+ \frac{39\beta SH_k}{T_{k,n_0}}\, ,
}
with probability at least $1-2\delta$. Using the lower bound $T_{k,n_0}\ge \frac{n^{1/4}}{4\sqrt{K}} $ yields
\als{
\sum_x G_k(x) \le \frac{T_{k,n_0}}{2\beta}b_{k,n_0+1} \le \sum_x G_k(x)
+ \frac{20\beta^{1/2}}{n^{1/8}}\sqrt{SKH_k \sum_x G_k(x)}+ \frac{156\beta SH_k\sqrt{K}}{n^{1/4}}\, ,
}
with probability at least $1-2\delta$.
Let us write the last inequality as
\als{
\sum_x G_k(x) \le \frac{T_{k,n_0}}{2\beta}b_{k,n_0+1} \le \sum_x G_k(x) + \epsilon_n\, ,
}
where $\epsilon_n = \widetilde\cO(n^{-1/8})$, thus giving
\als{
\frac{2\beta}{T_{k,n_0}}\sum_x G_k(x) \le b_{k,n_0+1} \le \frac{2\beta}{T_{k,n_0}}\Big(\sum_x G_k(x) + \epsilon_n\Big)\, .
}

\paragraph{Step 2: The second sub-problem.} 
Now let us consider $n_0 +1 \le t\le n$.
It follows that with probability $1-2\delta$, \BAMC\ allocates according to the following problem
\als{
\max_{\xi\in [0,\epsilon_n]^K}\max_{k} \frac{2\beta}{x_k + T_{k,n_0}}\Big(\sum_x G_k(x) + \xi_k\Big) \quad \text{s.t.:}\quad \sum_{k} x_k = n-\sqrt{n}\, ,
}
whose optimal solution satisfies
$$
\frac{(n-\sqrt{n})\sum_x G_k(x)}{\Lambda + K\epsilon_n} \le x_k \le \frac{(n-\sqrt{n})\Big(\sum_x G_k(x) + \epsilon_n\Big)}{\Lambda} \, .
$$
Recalling $\epsilon_n = \widetilde \cO(n^{-1/8})$ and noting that $T_{k,n}\ge x_k$, we obtain $\frac{T_{k,n}}{n} \to_{n\to\infty} \frac{\sum_x G_k(x)}{\Lambda}$.

It remains to show that the loss $L_{n}(\BAMC)$ approaches $\Lambda/n$ as $n$ tends to infinity. %To this end, consider a chain $k$, and let us denote by $\widetilde P_{k,n}$ the corresponding empirical estimator, that is for all $(x,y)\in \cS^2$, $\widetilde P_{k,n}(x,y) = \frac{1}{T_{k,x,n}}\sum_{t=1}^{n-1} \indic\{X_{t-1} = x, X_{t}=y\}$. Further, let $\widetilde L_{k,n}$ denote the corresponding loss for chain $k$.
%Using the same steps as in proof of Lemma \ref{lem:oracle_loss}, We have:
%\als{
%T_{k,n}\widetilde L_{k,n}\le T_{k,n}L_{k,n} \le T_{k,n}\widetilde L_{k,n} + \alpha^2 S^3\sum_x \frac{1}{T_{k,x,n}} + 2\sqrt{ \sum_x \frac{\alpha^2 S^3}{T_{k,x,n}} } \sqrt{T_{k,n}\widetilde L_{k,n}} \, .
%}
%And in particular, for $n\ge \text{\ncut}$, we have with probability at least $1-\delta$,
%\als{
%T_{k,n}\widetilde L_{k,n}\le T_{k,n}L_{k,n} \le T_{k,n}\widetilde L_{k,n} + \frac{2\alpha^2 S^3H_k}{T_{k,n}} + 2\sqrt{2\alpha^2 S^3\widetilde L_{k,n}} \, .
%\, .
%}
%Noting that we chose $\alpha = \frac{1}{3S}$, we get
%$$
%L_{k,n} \le + \widetilde L_{k,n}  + \frac{S}{9T_{k,n}^2} \, .
%$$
By Lemma \ref{lem:oracle_loss}, recall that $\lim_{T_{k,n}\to\infty} T_{k,n}L_{k,n} = \sum_x G_k(x)$. So, using  $\frac{T_{k,n}}{n} \to_{n\to\infty} \frac{\sum_x G_k(x)}{\Lambda}$, we conclude
$\lim_{n\to\infty} nL_{k,n} = \Lambda$, and the claim follows.
%Now consider the case when $n$ tends to infinity. Recall that $\frac{T_{k,n}}{n} \to_{n\to\infty} \frac{\sum_x G_k(x)}{\Lambda}$.
%Note that for this choice of $T_{k,n}$, Lemma \ref{lem:oracle_loss} implies that $n\widetilde L_{k,n} \to_{n\to\infty} \Lambda$. Thus,
%$
%nL_{k,n} \le n\widetilde L_{k,n} + \cO(n^{-1}),
%$
\ep

\end{document}